\documentclass[letterpaper]{article} 
\usepackage{aaai24}  
\usepackage{times}  
\usepackage{helvet}  
\usepackage{courier}  
\usepackage[hyphens]{url}  
\usepackage{graphicx} 
\usepackage{natbib}  
\usepackage{caption} 
\usepackage{algorithm}

\usepackage{booktabs}
\usepackage{multirow}
\usepackage{algpseudocode}
\usepackage{subcaption}
\usepackage{xcolor}
\usepackage{enumitem}
\usepackage{epsfig}
\usepackage{amsmath}
\usepackage{amssymb}
\usepackage{pifont}
\usepackage{eso-pic}
\usepackage{relsize}

\usepackage{newfloat}
\usepackage{listings}
\DeclareCaptionStyle{ruled}{labelfont=normalfont,labelsep=colon,strut=off} 
\floatstyle{ruled}
\newfloat{listing}{tb}{lst}{}
\floatname{listing}{Listing}
\title{EPSD: Early Pruning with Self-Distillation for Efficient Model Compression}
\author{
    Dong Chen\textsuperscript{\rm 1,2}\thanks{These authors contributed equally. This work was done during Dong Chen’s internship at Midea Group.},
    Ning Liu\textsuperscript{\rm 2}\footnotemark[1],
    Yichen Zhu\textsuperscript{\rm 2},
    Zhengping Che\textsuperscript{\rm 2},
    Rui Ma\textsuperscript{\rm 1}\thanks{Corresponding authors.},\\
    Fachao Zhang\textsuperscript{\rm 2},
    Xiaofeng Mou\textsuperscript{\rm 2},
    Yi Chang\textsuperscript{\rm 1},
    Jian Tang\textsuperscript{\rm 2}\footnotemark[2]
}
\affiliations{
    \textsuperscript{\rm 1}School of Artificial Intelligence, Jilin University\\
    \textsuperscript{\rm 2}Midea Group\\
    chendong21@mails.jlu.edu.cn,
    \{ruim, yichang\}@jlu.edu.cn,\\
    \{liuning22, zhuyc25, chezp, zhangfc, mouxf, jiantang22\}@midea.com

}

\begin{document}

\maketitle

\begin{abstract}
Neural network compression techniques, such as knowledge distillation (KD) and network pruning, have received increasing attention. Recent work `Prune, then Distill' reveals that a pruned student-friendly teacher network can benefit the performance of KD. However, the conventional teacher-student pipeline, which entails cumbersome pre-training of the teacher and complicated compression steps, makes pruning with KD less efficient. In addition to compressing models, recent compression techniques also emphasize the aspect of efficiency. Early pruning demands significantly less computational cost in comparison to the conventional pruning methods as it does not require a large pre-trained model. Likewise, a special case of KD, known as self-distillation (SD), is more efficient since it requires no pre-training or student-teacher pair selection. This inspires us to collaborate early pruning with SD for efficient model compression. In this work, we propose the framework named Early Pruning with Self-Distillation (EPSD), which identifies and preserves distillable weights in early pruning for a given SD task. EPSD efficiently combines early pruning and self-distillation in a two-step process, maintaining the pruned network's trainability for compression. Instead of a simple combination of pruning and SD, EPSD enables the pruned network to favor SD by keeping more distillable weights before training to ensure better distillation of the pruned network. We demonstrated that EPSD improves the training of pruned networks, supported by visual and quantitative analyses. Our evaluation covered diverse benchmarks (CIFAR-10/100, Tiny-ImageNet, full ImageNet, CUB-200-2011, and Pascal VOC), with EPSD outperforming advanced pruning and SD techniques.
\end{abstract}

\section{Introduction}
\label{sec:intro}
\begin{figure}[!tb]
\centering
\includegraphics[width=0.92\linewidth]{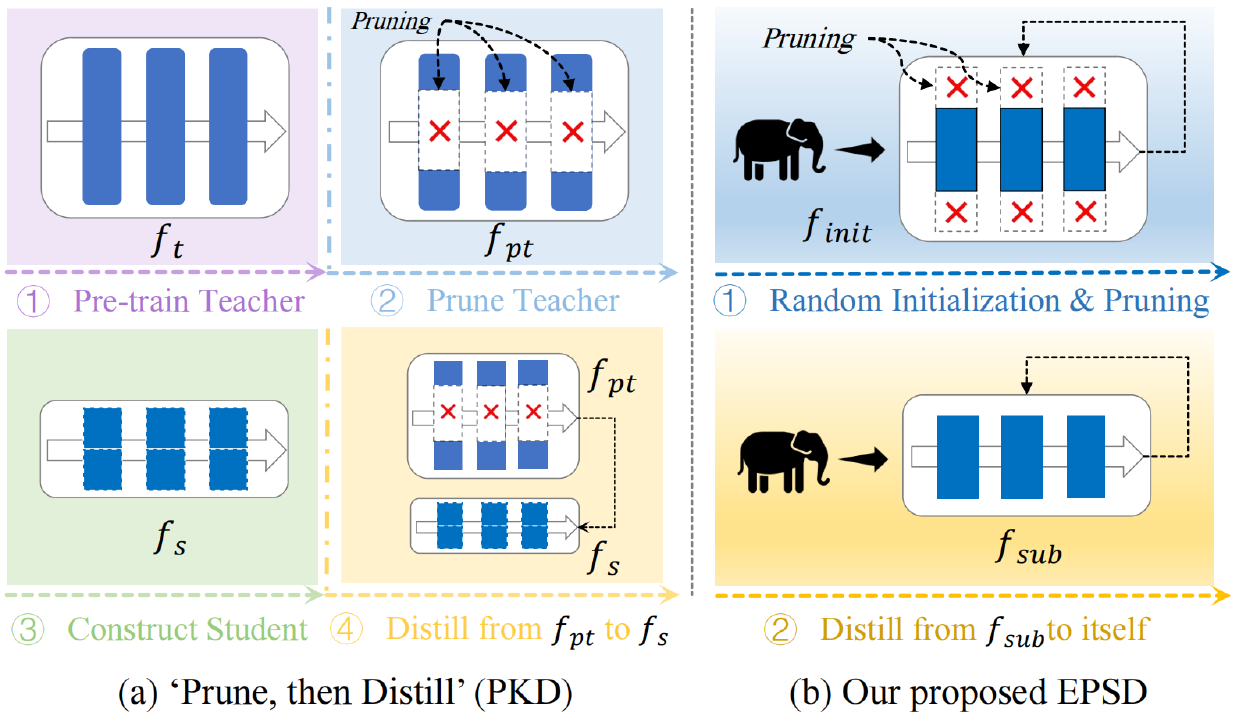}
\caption{Comparison of different model compression schemes. (a) PKD~\cite{park2021prune} follows four steps to combine pruning and KD. (b) Our Early Pruning with SD (EPSD) needs only two steps for compression.}
\label{fig:intro}
\end{figure}
%
Resource-limited edge devices struggle to handle the computational demands of large deep neural networks (DNNs). Therefore, compressing deep models is crucial to eliminate redundancy, facilitating the effective deployment of DNNs on edge devices~\cite{2019-iccv-dsq,liu2020autocompress,guo2023multidimensional}.
Various compression methods have been well studied, including network pruning~\cite{han2015learning,Guo_Ouyang_Xu_2020,Huang_2023_CVPR}, knowledge distillation (KD)~\cite{hinton2015distilling}, parameter quantization~\cite{hubara2017quantized,2022-neurips-outlier_suppression} and low-rank decomposition~\cite{zhang2015accelerating}.
Among them, KD and pruning have received increasing attention.

The concept behind KD is to train a smaller student network to approximate a larger, pre-trained teacher network with higher accuracy~\cite{hinton2015distilling}.
The cost of pre-training and the capacity gap issue between teachers and students inevitably limit the usage of KD~\cite{DBLP:conf/aaai/MirzadehFLLMG20,xu2019data}.
To overcome these limitations, self-distillation (SD) is proposed to enable students to distill knowledge from themselves~\cite{shen2022self,yang2019snapshot,zhang2021self,zhang2019your}.
Namely, SD allows the student network to learn from its predictions~\cite{mobahi2020self}, enabling a more streamlined training procedure that requires much fewer computational resources. 
However, the potential risk in SD is that the student could result in overfitting if the training process is not properly regularized~\cite{kim2021self}. 
Therefore, regularizing the student model becomes crucial to ensure effective knowledge transfer.

Network pruning removes the redundancy inside the original network and generates a sub-network with comparable accuracy performance~\cite{lecun1989optimal,liu2021lottery}.
In addition to reducing the computational requirements, pruning also helps prevent overfitting of DNNs~\cite{han2015deep}.
Current pruning works focus on pruning in the early stage~\cite{frankle2018lottery,de2020progressive,alizadeh2022prospect}. 
Pruning takes place either during initialization or shortly after a few training steps. Early pruning methods efficiently operate without the need for a pre-trained model.
Recent work `Prune, then Distill'~\cite{park2021prune} (we refer to it as PKD) explores the regularization effect of pruning on KD.
They analyze the distillation process regularized by the pruned teacher and combine pruning and KD in four steps as shown in Fig.~\ref{fig:intro}(a): 
1) Pre-train a teacher network $f_{t}$, 
2) prune $f_{t}$ and obtain a pruned teacher $f_{pt}$, 
3) construct a student network $f_{s}$ according to $f_{pt}$, and 
4) distill knowledge from $f_{pt}$ to $f_{s}$.
Though PKD reveals that \textit{pruning student-friendly teacher can boost the performance of KD}, the required cumbersome pre-training of the teacher and the complicated steps make it suffer from heavy training efforts.

To mitigate the complicated compression process, we attempt to collaborate early pruning with SD for efficient model compression. An intuitive approach is to prune the network and then finetune the pruned network with SD. 
However, different from PKD, in the context of the SD, pruning the teacher network also affects the student and leads to inadequate regularization if the pruned student presents weak trainability.
Empirically, applying a simple combination of pruning and SD results in severe performance degradation especially under the large sparsity ratios (as shown in Fig.~\ref{fig:degrade}). 
Therefore, the key question is: \textit{How to effectively prune DNNs with SD to yield performance gains?}

A promising solution is to make the pruned network favorable to SD, $i.e.,$ to preserve more \textit{distillable} weights to ensure the efficacy of the SD process.
To this end, we propose a novel framework named EPSD that collaborates \textbf{E}arly \textbf{P}runing and \textbf{S}elf-\textbf{D}istillation for efficient model compression. Specifically, EPSD has two main steps as shown in Fig.~\ref{fig:intro} (b):
1) \textbf{Early Pruning}: Prune an initialized network $f_{init}$ to obtain a pruned sub-network $f_{sub}$ with distillable weights. 
2) \textbf{Self-Distillation}: Train the pruned network $f_{sub}$ by SD.
Namely, given a desired sparsity level, EPSD globally ranks the weights in $f_{init}$ according to their influence (quantified by the absolute gradients) on the SD loss and removes weights with less influence.
By doing so, the trainability of the student network can be enhanced since the sub-network maintains objective consistency with SD loss and preserves more distillable weights.
Next, EPSD applies SD to recover the accuracy of the student network with distillable weights. 
Our contributions are summarized as follows:
\begin{itemize}[itemsep=1pt,topsep=1pt,parsep=1pt,leftmargin=12pt]
\item We present EPSD, which collaborates early pruning with SD, to compress models efficiently in only two steps. Meanwhile, EPSD preserves the trainability of the pruned network to improve performance.
\item EPSD identifies distillable weights that ensure objective consistency between early pruning and SD, and we present quantitative and visualized analysis to demonstrate the efficacy of EPSD.
\item Extensive results with three advanced SD methods on multiple benchmarks show that EPSD outperforms advanced pruning and SD methods while showcasing its scalability on two downstream tasks.
\end{itemize}

\section{Related Works}
\label{sec:related}
\noindent \textbf{Knowledge Distillation}.
Knowledge Distillation (KD) transfers various `knowledge' 
in networks~\cite{romero2014fitnets,hinton2015distilling}, acting as a potent regularization method to enhance generalization by utilizing learned softened targets~\cite{shen2022self}.
However, the capacity gap prevents well-performing teachers from making students better~\cite{DBLP:conf/aaai/MirzadehFLLMG20}.

\noindent \textbf{Self-Distillation}.
To improve the efficiency of knowledge transfer, Self-Distillation (SD) leverages knowledge from the student network without involving additional teachers~\cite{wang2021knowledge,yun2020regularizing}.
The key to SD is creating soft targets, where the student network generates its valuable knowledge to guide its training~\cite{lee2020self,zhang2021self,yang2019snapshot,shen2022self}.
SD's efficiency arises from avoiding teacher network pre-training and addressing teacher-student capacity gaps. Yet, the student network might be over-fitting due to insufficient training regularization~\cite{kim2021self}.
Recently, PKD~\cite{park2021prune} revealed the positive regularizing impact of pruning teacher networks on KD, which inspires us to regularize the SD process by pruning.

\noindent \textbf{Network Pruning}. 
Network pruning aims to identify and remove unnecessary weights, reducing complexity while preserving training performance~\cite{reed1993pruning, lee2019signal}.
Traditional approaches~\cite{han2015learning, molchanov2016pruning} typically follow pre-training, pruning, and re-training to prune, which requires much training effort.
Another paradigm named Dynamic Sparse Training (DST)~\cite{mocanu2018scalable,bellec2017deep,evci2020rigging,liu2021we} starts from a (random) sparse neural network and allows the sparse connectivity to evolve dynamically during training. DST can significantly improve the trainability of sparse DNNs without increasing the training FLOPs.
Recently, early pruning~\cite{lee2018snip,wang2020picking,de2020progressive,alizadeh2022prospect} has been widely studied as it identifies sparse sub-networks before training without cumbersome pre-training.
Many early pruning works evaluate the importance of individual weights regarding the impact on loss, i.e., the gradients of a network.
Though early pruning is efficient, it is considered under-performance~\cite{wang2022recent}: pruning neural networks breaks the dynamical isometry~\cite{saxe2013exact} and results in the trainability degradation~\cite{lee2019signal}. 
In this work, we empirically show that SD greatly enhances the performance of early pruned networks and improves their trainability by ensuring alignment between pruning and SD objectives.

\section{Early Pruning with Self-Distillation}
\label{sec:3-EPSD}
\begin{figure}[!t]
\centering
\includegraphics[width=1.0\linewidth]{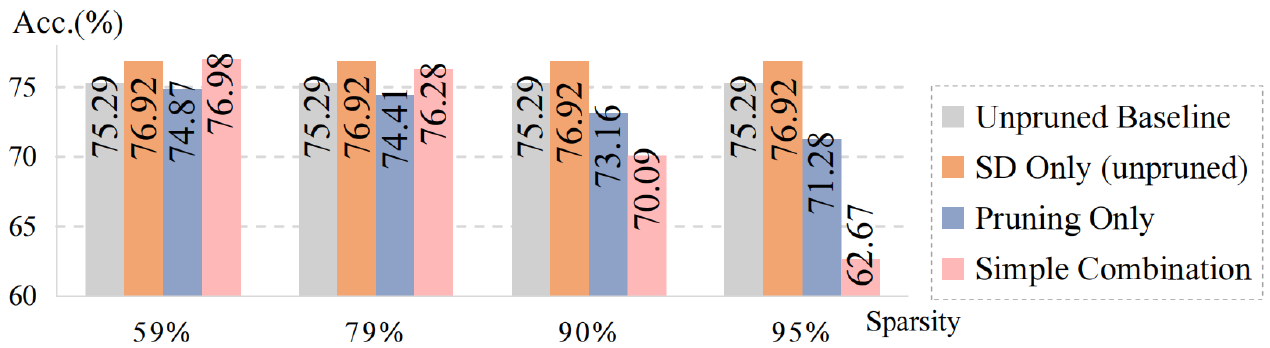} 
\caption{Performance comparison among the `Simple Combination', pre-trained network (`Unpruned Baseline'), the network only performs pruning without fine-tuning with SD (`Pruning Only'), and the network only performs SD without any sparsity (`SD Only') on CIFAR-100 of ResNet-18. The `Simple Combination' suffered severe performance degradation, especially under the high sparsity ratio $95\%$.}
\label{fig:degrade}
\end{figure}
\begin{figure*}[!t]
\centering
\includegraphics[width=0.95\linewidth]{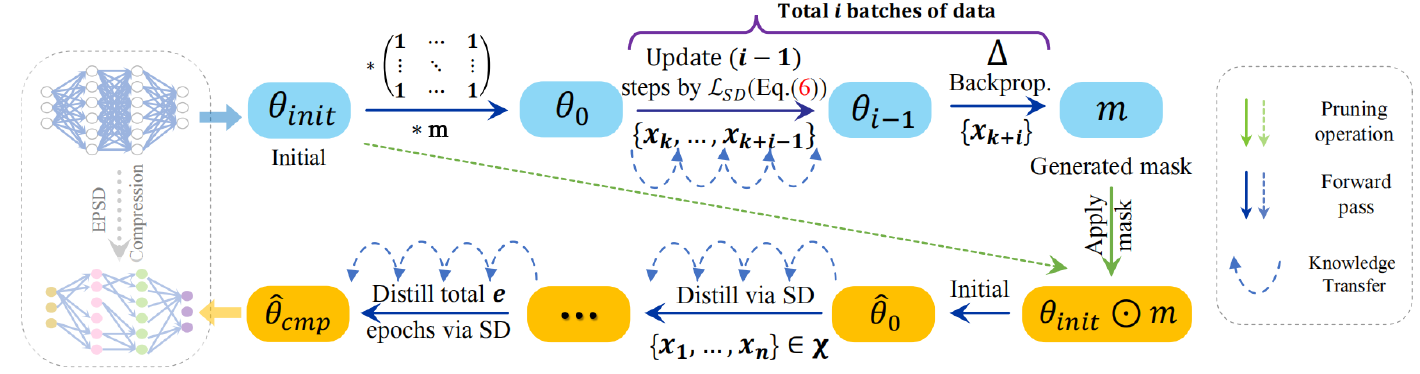}
\caption{
EPSD prunes a random initialized network with weights $\theta_{init}$ in step-1 (blue block) and then employs the SD algorithm to train the pruned network in step-2 (orange block). In Step 1, EPSD identifies and retains distillable weights by measuring the impact of SD loss on individual weights after $i$ steps of training.}
\label{fig:pipeline}
\end{figure*}
We first introduce a simple combination of early pruning and SD and show that it suffers performance degradation. To address this issue, we introduce the concept of distillable weights, along with quantitative and visualized analysis. Finally, we present the overall framework of EPSD, demonstrating the efficiency by comparing the required training efforts with other compression techniques.

\subsection{The `Simple Combination'}
\label{section:3_1}
A straightforward way to combine pruning and SD requires two steps. 
Step-1: Network pruning without pre-training and 
step-2: Distill knowledge to itself.
Specifically, step-1 is to identify a sub-network from the randomly initialized network by pruning. Step-2 is to fine-tune the sub-network via SD. 
Since our goal is to efficiently compress the model, the early pruning method ProsPr~\cite{alizadeh2022prospect} is utilized as the representative method in step-1.

\noindent \textbf{Step-1: Identify Redundancy Before Training}. 
Lee \textit{et al.} first proposed SNIP~\cite{lee2018snip} to prune unnecessary weights in random initialized networks that are least salient for the loss.
They compute the gradients $\Delta$ to generate saliency scores for initial weight $\theta_{init}$ with random samples $x_{rand}$ and remove the weights with the lowest scores.
Specifically, an all-one mask $m$ is attached to initial weights to get $\theta_{0} \gets m \odot \theta_{init}$, and the saliency scores can be computed as:
\begin{equation}
\label{equ:snip_1}
{\Delta}(w_{p}, x_{rand})=\frac{\partial \mathcal{L}(\theta_{0}, x_{rand})}{\partial m_{p}},
\end{equation}
\begin{equation}
\label{equ:snip_2}
s_{w_{p}}=\frac{\left|{\Delta}(w_{p}, x_{rand})\right|}{\sum_{q}\left|{\Delta}(w_{q}, x_{rand})\right|} ,
\end{equation}
where $m$ is the pruning mask with values 0s or 1s (initial value is $1$s), 
$\Delta$ denotes gradients derived from labels,
$w_{p}$ is $p$-th weight in $\theta_{init}$, 
$s_{w_{p}}$ is the saliency score for measuring the importance of $w_{p}$.
Recently, Milad et al. pointed out that pruning should consider the trainability of a certain weight, instead of only its immediate impact on the loss before training~\cite{alizadeh2022prospect}, they measured the impact of pruning on loss across $i$ gradient descent steps during initial training, rather than assessing alterations in loss at initialization.
The saliency scores are calculated based on the updated weights $\theta_{i}$ as in Eq.~\eqref{eq:prospr}:
\begin{equation}
\label{eq:prospr}
{\Delta}(w_{p}, x_{i})=\frac{\partial \mathcal{L}(\theta_{i}, x_{i})}{\partial m_{p}},
\end{equation}
where $x_{i}$ denotes $i$-\textit{th} random sampled batch of data for computing the gradients. 
In the classification tasks, the cross-entropy (CE) loss runs through the entire process, from pruning to training. The difference between predictions and labels is used to evaluate the importance of weights and optimize the pruned network.

\noindent \textbf{Step-2: Distilling Knowledge from Soften Targets}.
\label{section:3_1_2}
In classification task, we denote $\mathbf{x} \in \mathcal{X}$ as input and \(y \in \mathcal{Y}=\)\(\{1, \ldots, C\}\) as its ground-truth label.
Given the input $\mathbf{x}$, the predictive distribution of a softmax classifier is:
\begin{equation}
P(y \mid \mathbf{x} ; \theta, \tau)=\frac{\exp \left(l_{y}(\mathbf{x} ; \theta) / \tau\right)}{\sum_{i=1}^{C} \exp \left(l_{i}(\mathbf{x} ; \theta) / \tau\right)},
\end{equation}
where \(l_{i}\) denotes the logit of DNNs for class \(i\) which are  parameterized by \(\theta\), and \(\tau>0\) is the temperature scaling factor.
To improve the generalization ability, traditional KD~\cite{hinton2015distilling} transfers pre-trained teacher's knowledge by optimizing an additional Kullback-Leibler (KL) divergence loss between the softened outputs $\widetilde{P}$ from teacher and student in every mini-batch $x_{i}$:
\begin{equation}
\label{equ:kd_loss}
    \mathcal{L}_{K D}=\frac{1}{n} \sum_{i=1}^{n} \tau^{2} \cdot D_{K L}\left(\widetilde{P}({x_{i};\theta_{t}}) \| \widetilde{P}({x_{i};\theta_{s}})\right).
\end{equation}

The original KD matches the predictions of the same inputs from two different networks, while the SD replaces the teacher's prediction with that of the student network itself.
Various works~\cite{lee2020self,xu2019data,zhang2019your,zhang2021self,yang2019snapshot,shen2022self} have explored enhancing the SD method in different ways. 
Our work focuses on the gradients of SD loss rather than specific improvements.  
We further discuss the gradients of SD loss in Sec.~\ref{section:3_2}.
Without loss of generality, we formulate SD loss as follows:
\begin{equation}
\label{equ:sd_loss}
    \mathcal{L}_{S D}=\frac{1}{n} \sum_{i=1}^{n} \tau^{2} \cdot D_{K L}\left(\widetilde{P}({\overline{x}_{i};\overline{\theta}_{s}}) \| \widetilde{P}({x_{i};\theta_{s}})\right),
\end{equation}
where $\widetilde{P}({\overline{x}_{i};\overline{\theta}_{s}})$ represents the soft targets produced by the student networks in SD, and different SD methods have different definitions of $\overline{x}_{i}$ and $\overline{\theta}_{s}$. 
We refer the readers to the Appendix for a more detailed explanation of these symbols.

\noindent \textbf{Remarks}. 
A baseline method for combing early pruning and SD involves applying the two techniques sequentially (we name it `Simple Combination.').
This straightforward approach produces a distilled, sparse network.
The preliminary study shown in Fig.~\ref{fig:degrade} demonstrates that under the sparsity ratio of 95\%, the accuracy of the `Simple Combination' is only 62.67\%,  lower nearly 13\% than the `Unpruned Baseline' and 8\% than the `Pruning Only'. 
These anomalous results indicate that the pruned network can not effectively learn via SD when directly applying SD to the early-pruned network. 
To this end, we raised one question: \textit{``Why does the early-pruned network degrade accuracy when training with SD?"}
In the pruning step of the `Simple Combination', the gradient only reflects the difference between the network output and the hard labels, without considering the soft targets generated in SD.
We argue that it is difficult for the early-pruned network to learn knowledge from itself during SD when directly combining the early pruning and SD.

\subsection{Identify Distillable Weights via SD}
\label{section:3_2}
As introduced in the previous section, a simple combination of early pruning and SD does not lead to performance gains and even results in severe degradation at large sparsity. 
In the SD scenario, when the teacher network is pruned, the students are also affected, potentially leading to inadequate distillation if a weak student is involved.
A desirable mitigation solution is to make pruning results favorable to SD, $i.e.,$ to \textit{preserve more distillable weights} to ensure that the pruned model can be better distilled. Intuitively, obtaining distillable weights implies that the pruned network should be consistent with the optimized objective of SD.

As a result, we propose to identify distillable weights with SD loss before training.
More specifically, during pruning, we establish a knowledge transmission path to facilitate the model to learn from its own outputs.
We evaluate the importance of the weights by conducting a few SD iterations to derive the necessary gradients.
Formally, the salience score for an individual weight can be derived from Eq.~(\ref{eq:prospr}) and~(\ref{equ:sd_loss}):
\begin{equation}
\label{eq:soft_gradient}
{\widetilde{\Delta}}(w_{p}, x_{i})=\frac{\partial \mathcal{L}_{SD}(\theta_{i}, x_{i})}{\partial m_{p}},
\end{equation}
\begin{equation}
\label{eq:saliency_score}
\widetilde{s}_{w_{p}}=\frac{\left|{\widetilde{\Delta}}(w_{p}, x_{i})\right|}{\sum_{q}\left|{\widetilde{\Delta}}(w_{q}, x_{i})\right|}.
\end{equation}
We remove weights that have the least impact on SD loss according to the desired sparsity ratio, and the \textit{weights with higher salience scores $\widetilde{s}$ are regarded as distillable to be preserved}.
We thoroughly assess weight importance by considering both hard label influences and network-generated knowledge during pruning, resulting in more reliable saliency criteria driven mainly by the SD loss.

\begin{figure}[ht!]
\centering
\begin{subfigure}{0.48\textwidth}
\centering
\includegraphics[width=\textwidth]{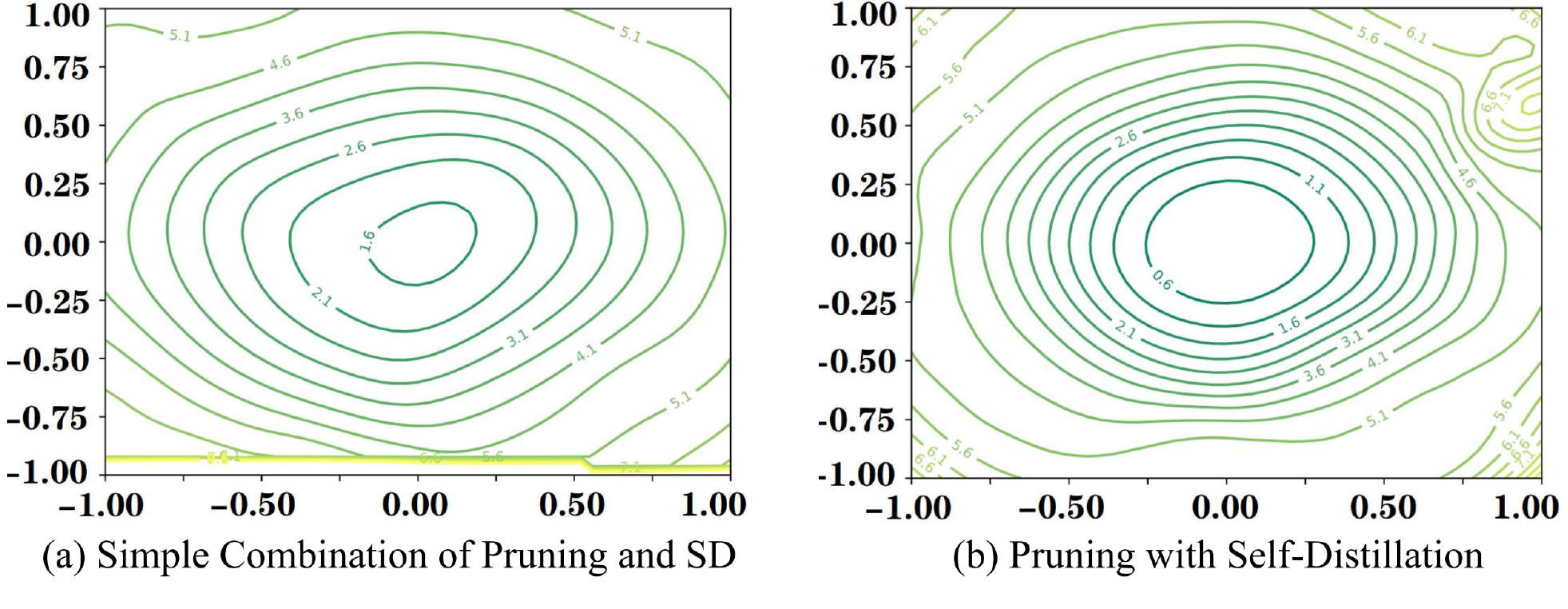}
\end{subfigure}
\begin{subfigure}{0.48\textwidth}
\centering
\includegraphics[width=\textwidth]{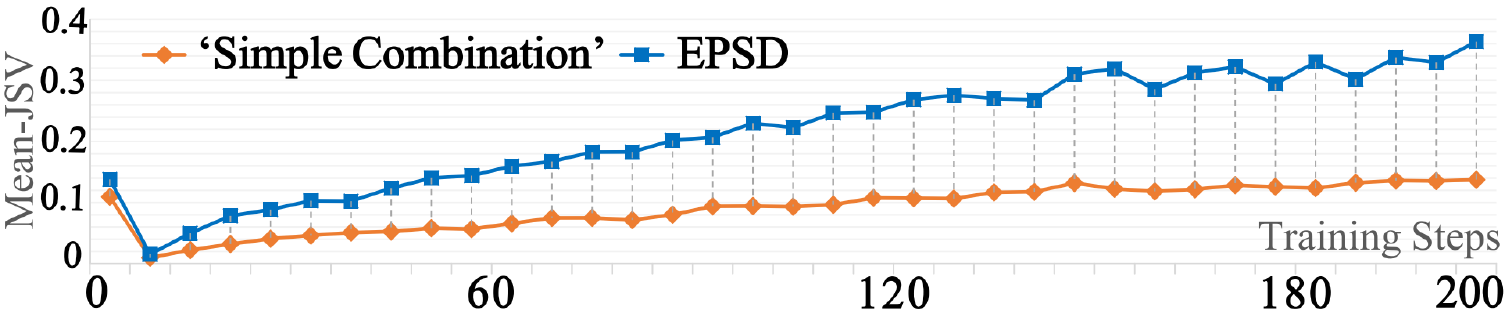}
\end{subfigure}
\caption{Trainability analysis. Top: Loss contour plots of early-pruned networks using (a) `Simple Combination' and (b) EPSD. Bottom: Comparison of Mean-JSV curves of EPSD and the `Simple Combination' approach.}
\label{fig:trainability} 
\end{figure}

To analyze the trainability of the pruned model, we leverage loss surface~\cite{li2018visualizing} to visualize the loss landscape and assess the ease of optimization. Additionally, we utilize the mean Jacobian singular values (Mean-JSV) as a quantitative metric to gauge compliance with the dynamic isometry conditions~\cite{wang2021dynamical, DBLP:conf/iclr/Wang023a}.
The top of Fig.~\ref{fig:trainability} shows the contour plots of loss. We observed that the loss surface of EPSD is flatter than the `Simple Combination', and reaches local minima faster (minimum loss value $0.6$ v.s $1.6$ within equal training steps), implying the pruned model by EPSD is easier to optimize~\cite{arora2018stronger, dinh2017sharp}.
The bottom of Fig.~\ref{fig:trainability} shows Mean-JSV curves over the first 200 training steps for pruned model obtained by EPSD and the `Simple Combination', respectively.
In theory, a larger Mean-JSV (closer to 1) indicates better trainability of the model. The Mean-JSV of EPSD better meets dynamic isometry requirements than the `Simple Combination', revealing the potential of keeping objective consistency between pruning and SD in preserving trainable weights.

\noindent \textbf{Remarks}. 
To tackle the degradation issue raised by the `Simple Combination', we aim to pinpoint distillable weights preferred by SD for improved accuracy. Our visual and quantitative analysis reveals that sub-networks identified by maintaining objective consistency exhibit superior trainability compared to those identified solely through pruning.

\subsection{Towards Efficient Model Compression}
\label{section:3_3}
\begin{figure}[!tb]
\centering
\includegraphics[width=1\linewidth]{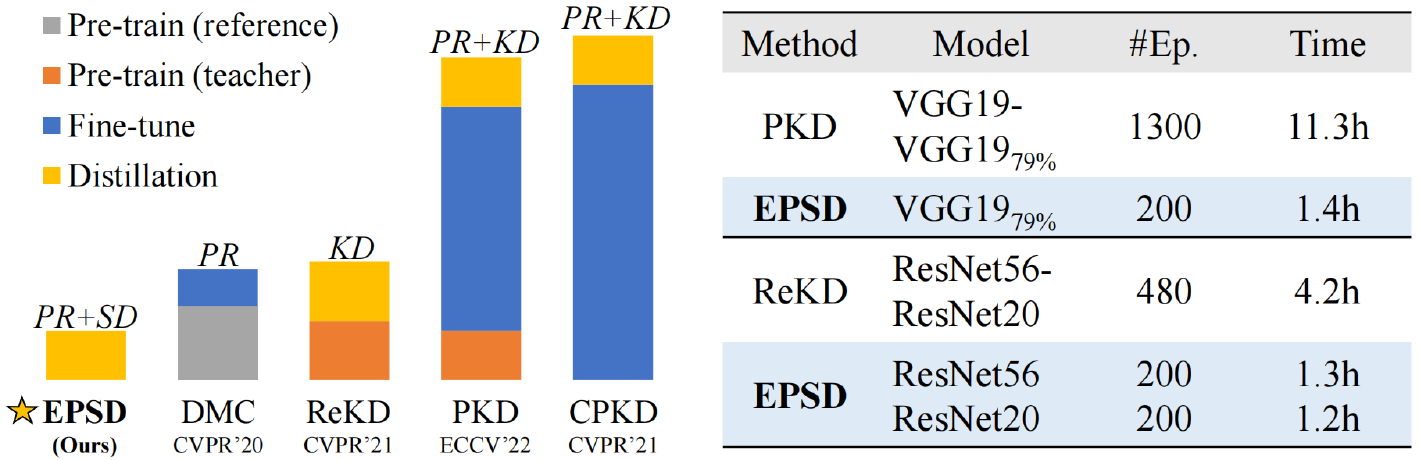} 
\caption{
Training efforts comparisons among various representative compression approaches. 
Left: Total training epochs of CPKD~\cite{aghli2021combining}, PKD~\cite{park2021prune}, ReKD~\cite{chen2021distilling}, DMC~\cite{gao2020discrete}. `PR' and `KD (SD)' denote pruning and knowledge distillation (self-distillation), respectively. Right: Comparison of total training wall time under identical conditions.}
\label{fig:efforts} 
\end{figure}
Fig.~\ref{fig:pipeline} shows the overall compression procedure of EPSD.
There are mainly two steps as mentioned in Sec.~\ref{section:3_1}. 
In step-1, given randomly initialized weights $\theta_{init}$, EPSD estimates the effect of pruning on the SD loss (Eq.~\ref{equ:sd_loss}) over $i$ steps of gradient descent.
By doing so, EPSD preserves more distillable weights, which become crucial since they offer superior trainability and are more easily optimized by the SD loss as discussed in Sec.~\ref{section:3_2}.
Once the pruning mask $m$ is generated by the gradients $\widetilde{\Delta}$, we apply it to the initial weights $\theta_{init}$ to get a pruned network.
In Step 2, we train the pruned network by SD until it reaches convergence.

We emphasize that EPSD is efficient, which is attributed to: 1) the absence of pre-training for pruning, 2) the elimination of teacher training, and 3) the pruned network's distillable weights, which contribute to improved trainability and faster convergence during SD.
In Fig.~\ref{fig:efforts}, we demonstrate the training efforts of EPSD and compare them against other representative compression methods. Among them, EPSD combines early pruning and SD (\textit{PR+SD}), DMC uses advanced pruning (\textit{PR}), ReKD is a KD method (\textit{KD}), and the other two are combinations of pruning and KD (\textit{PR+KD}). 
EPSD achieves efficient training with fewer epochs than other methods. 
For instance, the training time of PKD is about eight times that of EPSD (11.3 vs. 1.4 hours).

\section{Experiments}
\label{sec:exps}
\begin{figure*}[!ht]
\centering
\begin{tabular}{ccc}
    \includegraphics[width=0.32\textwidth]{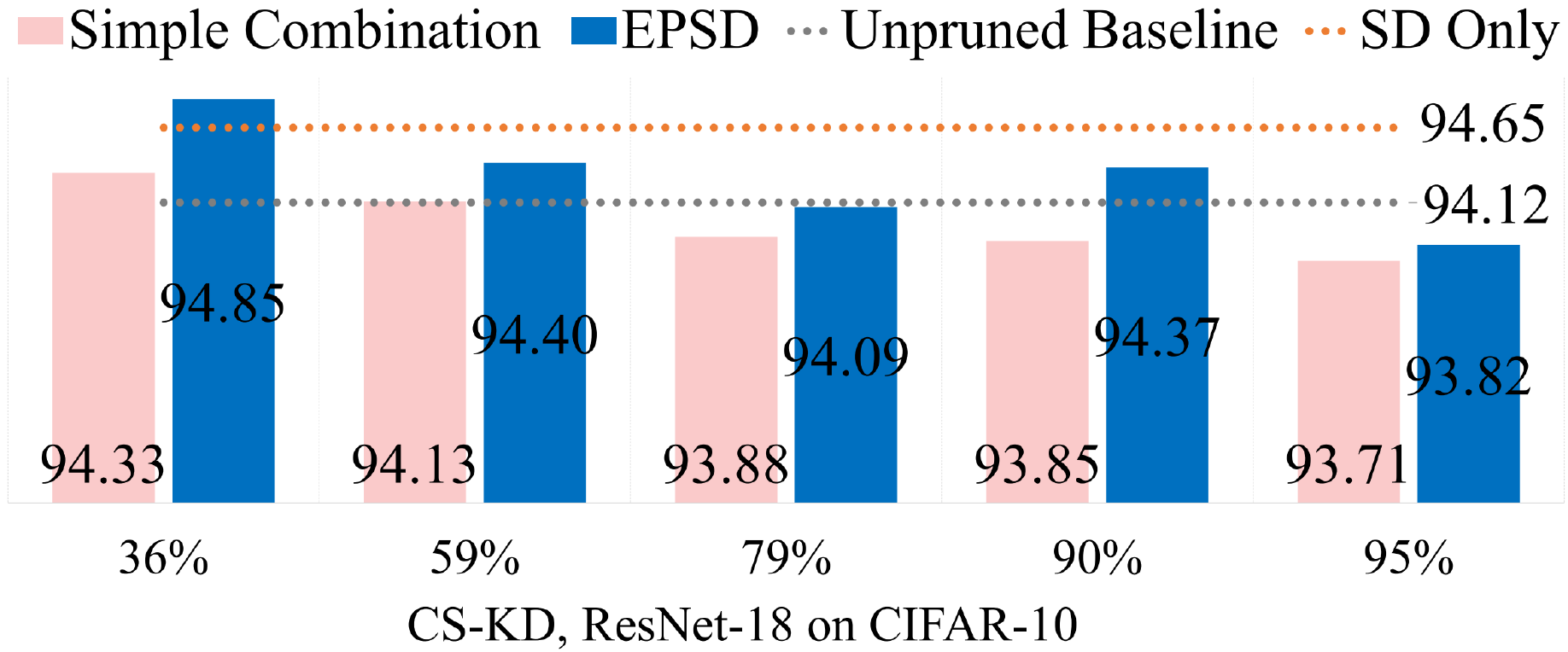} \hspace{-4.2mm} 
    & \includegraphics[width=0.32\textwidth]{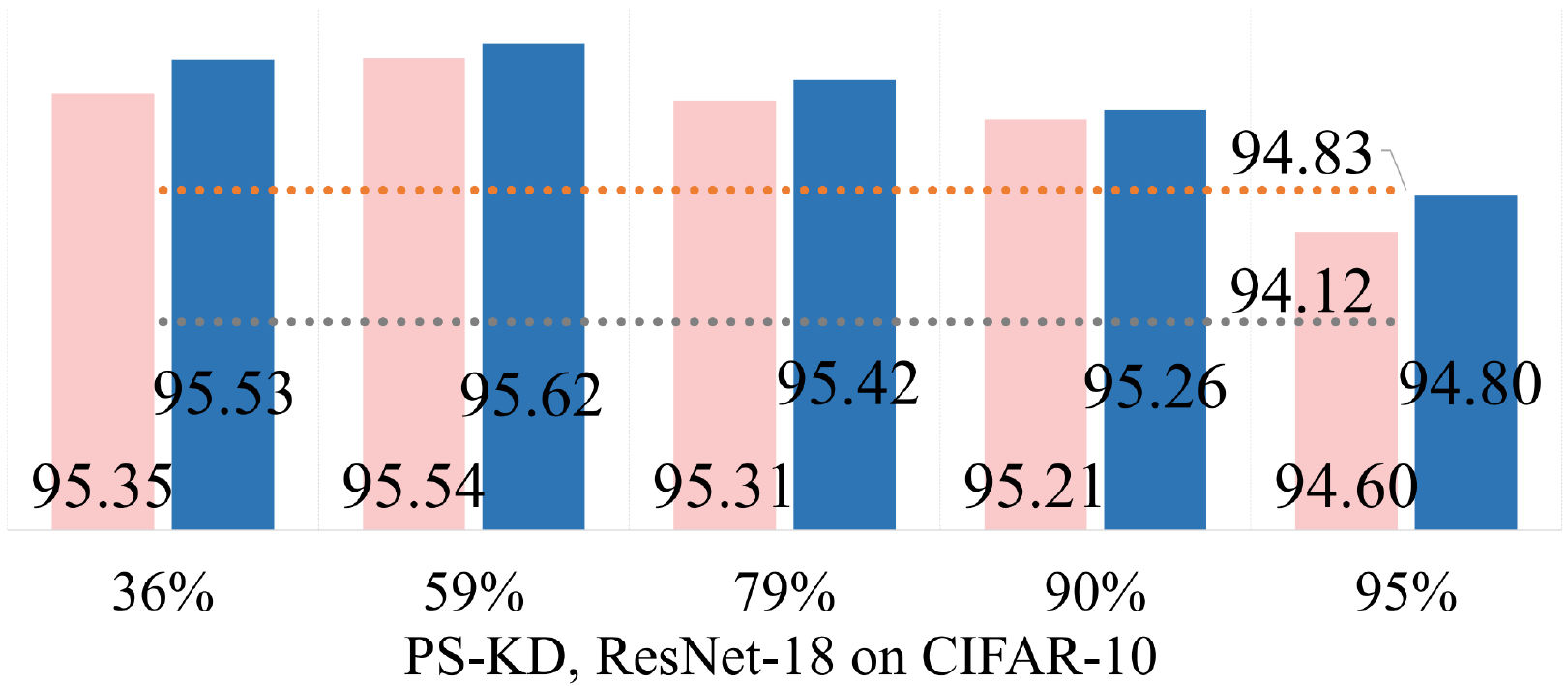} \hspace{-4.2mm}
    & \includegraphics[width=0.32\textwidth]{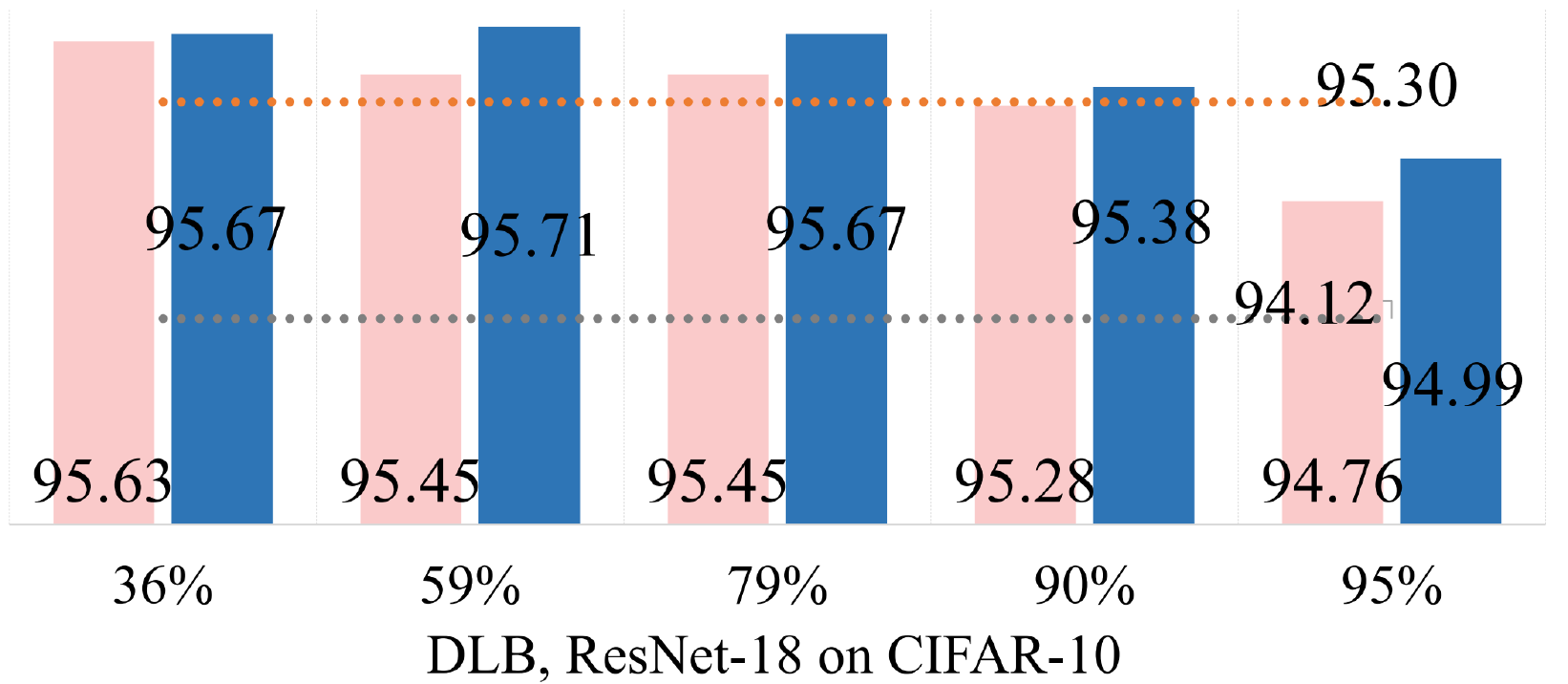} \\
    
    \includegraphics[width=0.32\textwidth]{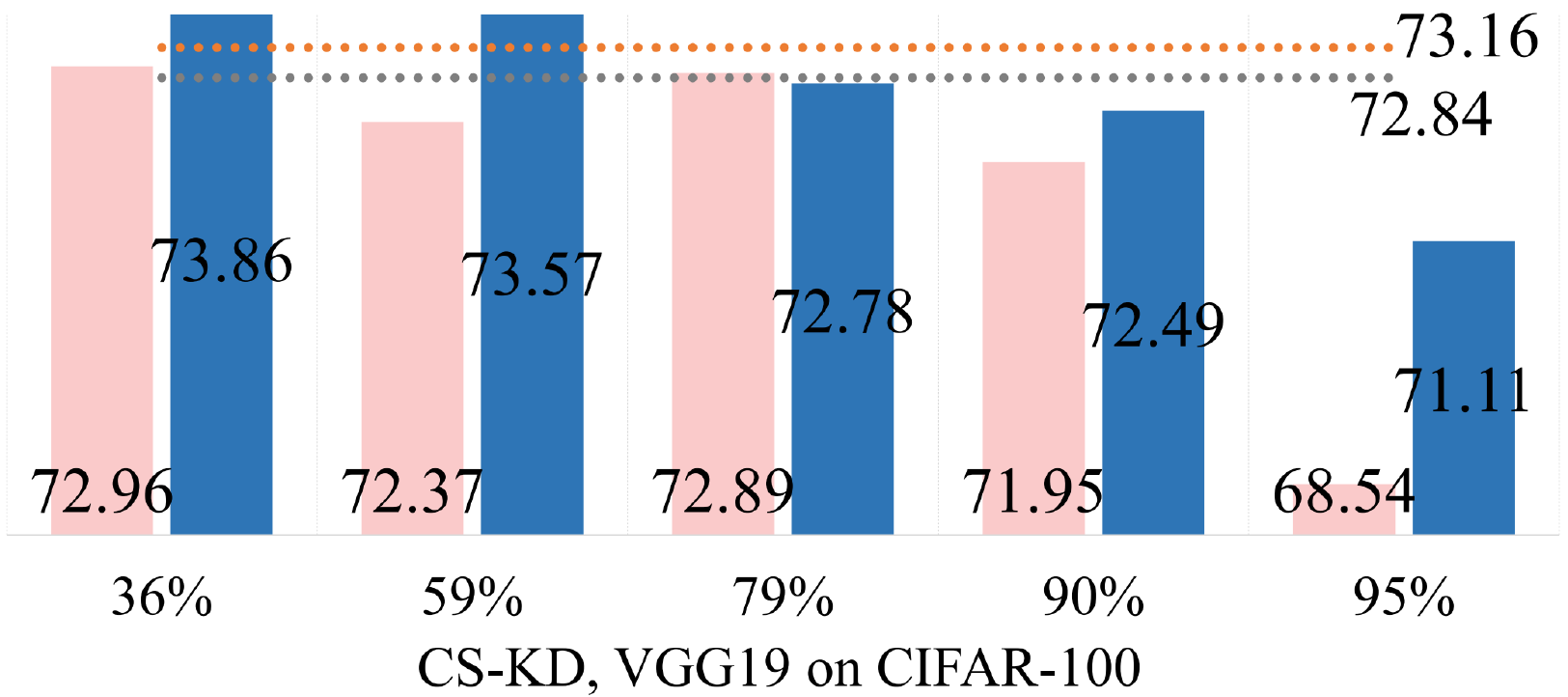} \hspace{-4.2mm}
    & \includegraphics[width=0.32\textwidth]{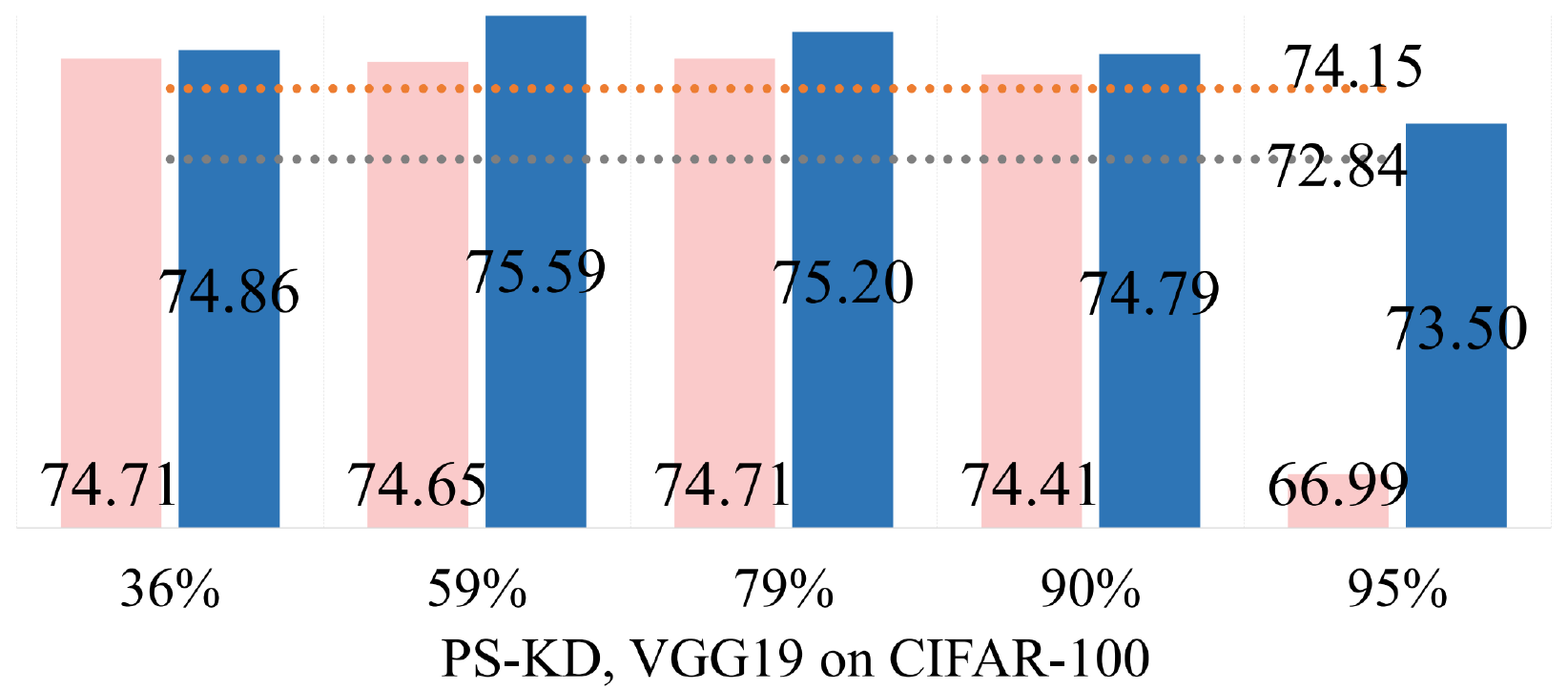} \hspace{-4.2mm}
    & \includegraphics[width=0.32\textwidth]{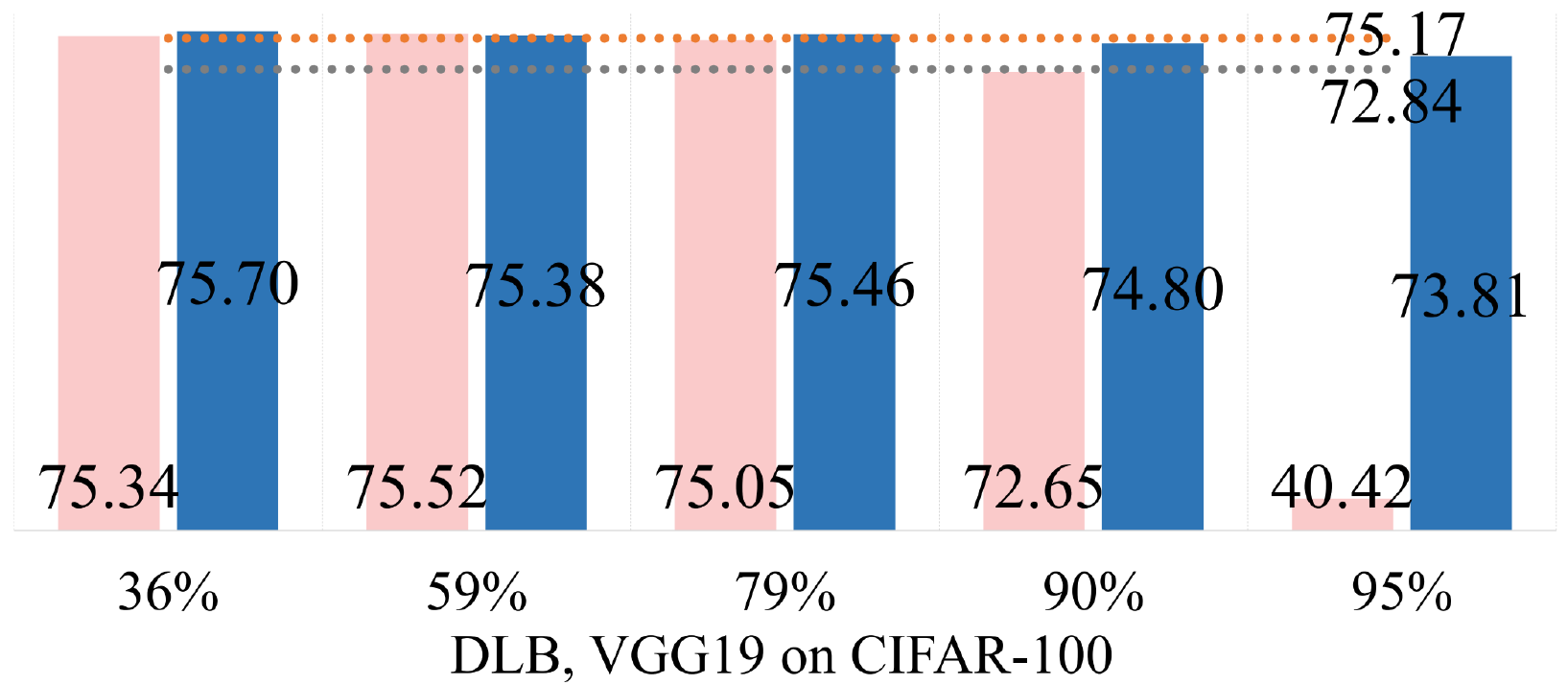} \\
    
    \includegraphics[width=0.32\textwidth]{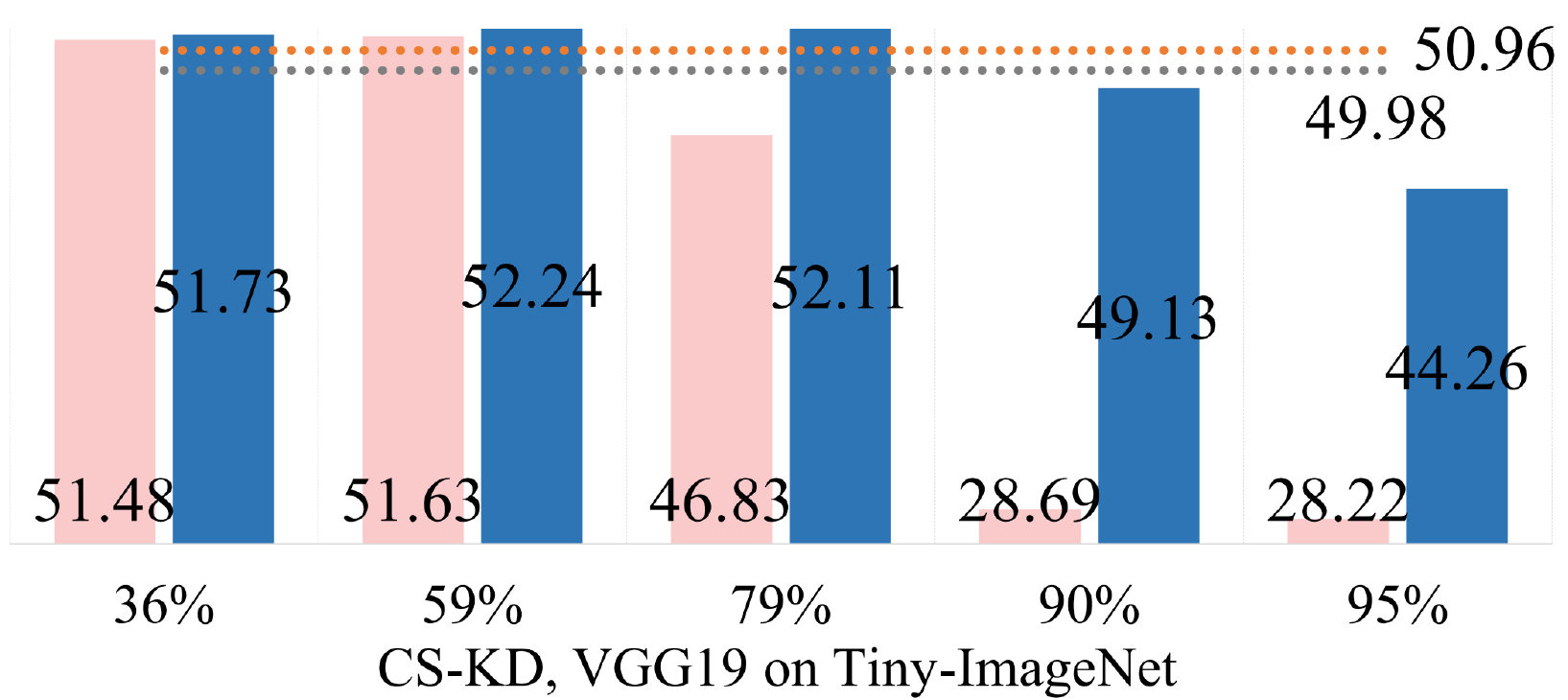} \hspace{-4.2mm}
    & \includegraphics[width=0.32\textwidth]{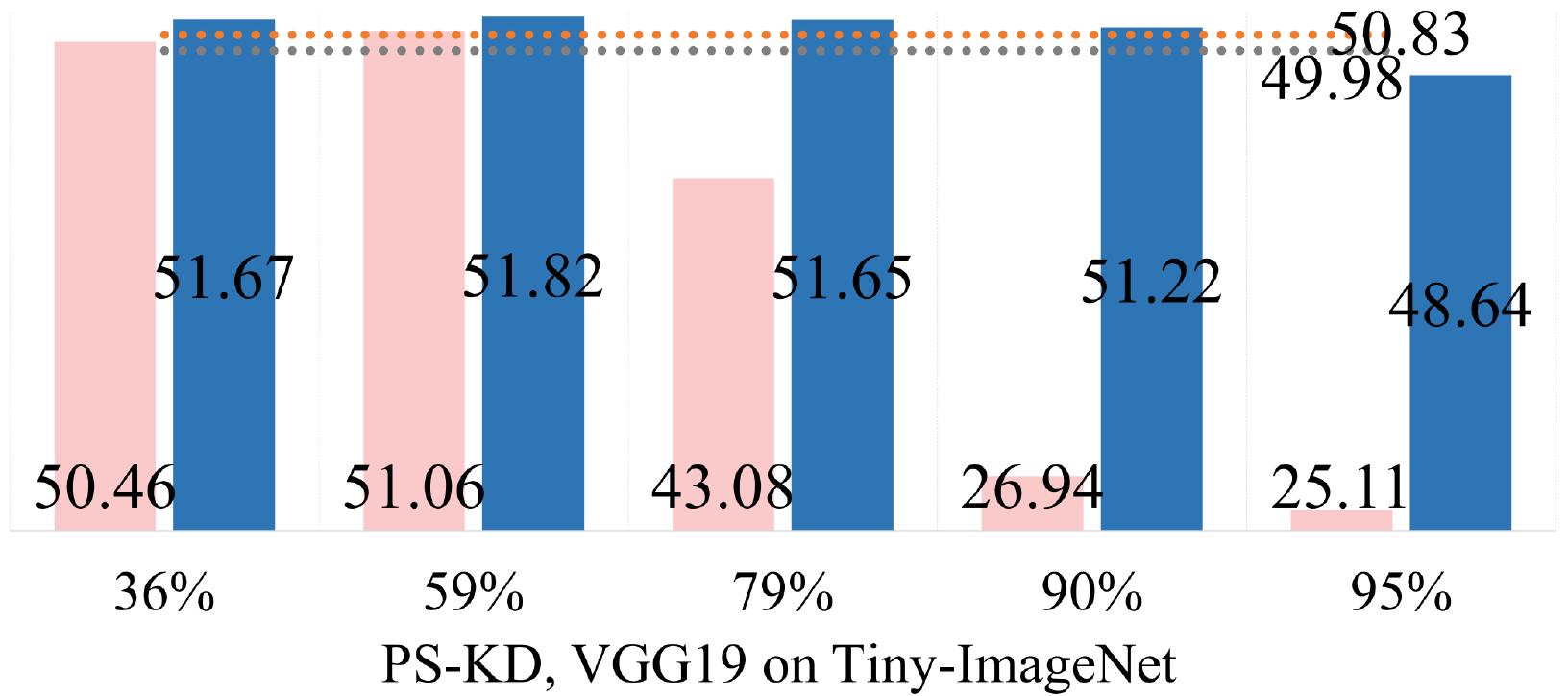} \hspace{-4.2mm}
    & \includegraphics[width=0.32\textwidth]{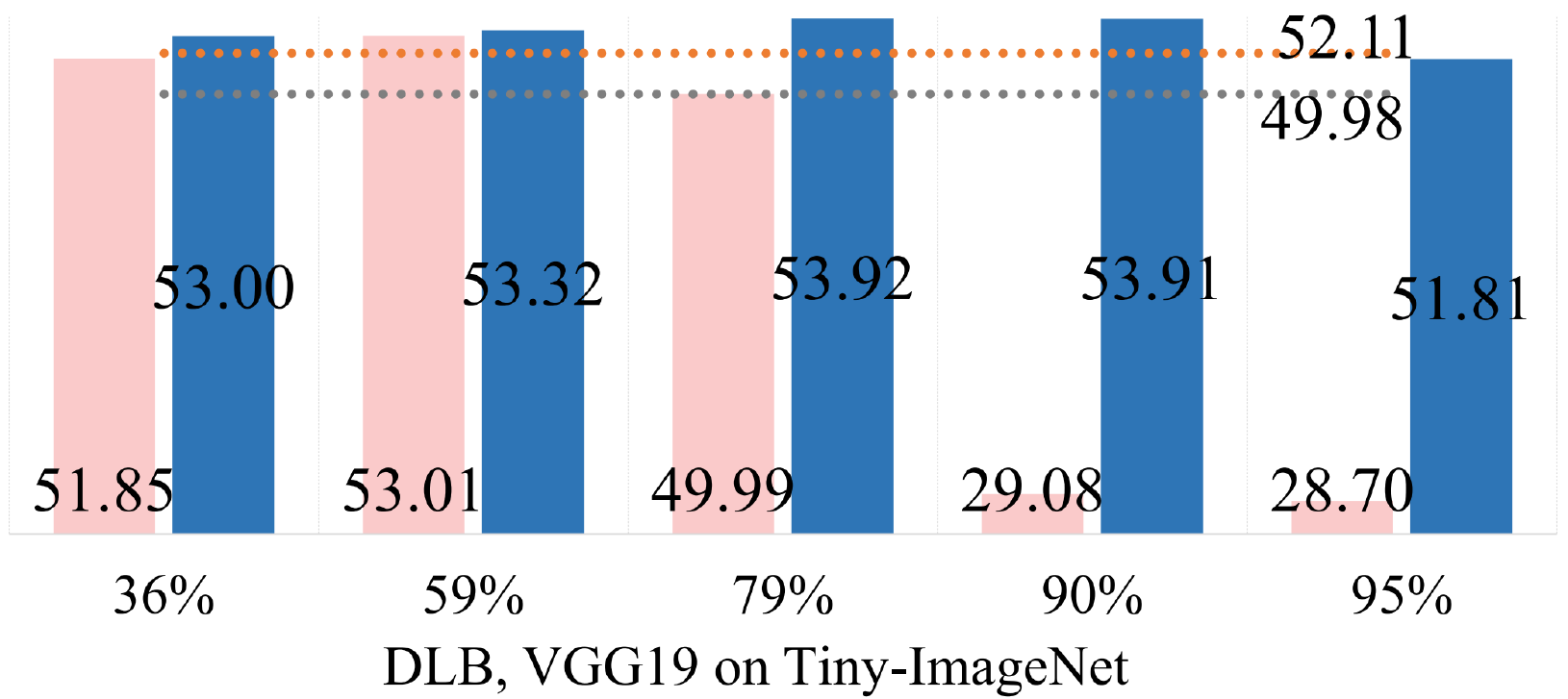} \\
\end{tabular}
\caption{Test accuracy among `Unpruned Baseline', `SD Only', `Simple Combination' and EPSD with three equipped SD methods (CS-KD, PS-KD, DLB) on CIFAR-10/100 and Tiny-ImageNet under various sparsity.
The three rows illustrate the three datasets and the three columns display the three equipped SD methods.}
\label{fig:main_results}
\end{figure*}
We evaluate EPSD on various benchmarks, including CIFAR-10/CIFAR-100~\cite{krizhevsky2009learning}, Tiny-ImageNet, and full ImageNet~\cite{deng2009imagenet} using diverse networks and comparing with the `Simple Combination' approach, advanced pruning and SD methods.
We also assess EPSD's adaptability and scalability in two downstream tasks.
More details can be found in the Appendix.

\subsection{EPSD equipped with Various SD Methods}
\label{sec:4.1}
We incorporate three distinct SD algorithms (CS-KD~\cite{yun2020regularizing}, PS-KD~\cite{kim2021self}, and DLB~\cite{shen2022self}) into EPSD to ensure a comprehensive evaluation. Our experiments are conducted on CIFAR-10/100 and Tiny-ImageNet datasets across five sparsity ratios (36\%, 59\%, 79\%, 90\%, 95\%). To ensure fairness in comparison, we employ identical hyper-parameters for training each dataset. For each variant of EPSD utilizing a specific SD method, we conduct a comprehensive comparison with 1) the unpruned network without any pruning or SD (Unpruned Baselines'), 2) network training using the respective SD method ('SD Only'), and 3) the simple combination of pruning and the specific SD method (`Simple Combination').
Figure~\ref{fig:main_results} illustrates the specific comparison results.

Based on the results, we have the following observations:
\begin{itemize}[itemsep=1pt,topsep=1pt,parsep=1pt,leftmargin=12pt]
\item 
  EPSD consistently outperformed the `Simple Combination' overall settings. 
  Moreover, under high sparsity conditions ($e.g.,$ $95\%$), EPSD remained competitive while the `Simple Combination' heavily declined.
\item
  On the more challenging Tiny-ImageNet, the `Simple Combination' degraded more severely than EPSD for all three SD methods.
  For instance, with DLB and VGG-19 on Tiny-ImageNet at sparsity $90\%$, the accuracy of the `Simple Combination' is $20.80\%$ lower than `Unpruned Baseline' ($29.08\%$ vs. $49.88\%$),
  while EPSD achieved $53.91\%$ accuracy, increasing $1.80\%$ and $3.93\%$ compared to `SD Only' and `Unpruned Baseline', respectively.
\item
  EPSD outperformed `Unpruned Baseline' and `SD Only' over all three SD methods in most settings, indicating that early pruning with SD can boost the performance of SD.
  EPSD maintains an advantage over the `Simple Combination', affirming its efficacy in preserving more distillable weights and achieving promising performance.
\end{itemize}

\subsection{Comparison of Pruning Methods}
\label{sec:4.2}
To illustrate the effectiveness of EPSD, we compared EPSD with advanced pruning methods on CIFAR-10/100 (See Appendix) and ImageNet. 
Further, we extended EPSD with structured pruning to show the extensibility of our method.

\begin{table}[!tb]
\centering
\small
\renewcommand\arraystretch{1.05} 
\setlength{\tabcolsep}{1.2mm}{
\begin{tabular}{ccccccccc}
\toprule
Backbone & \multicolumn{4}{c}{VGG-19} & \multicolumn{4}{c}{ResNet-50} \\ \midrule
Sparsity & \multicolumn{2}{c}{ 90\% } & \multicolumn{2}{c}{ 95\% } & \multicolumn{2}{c}{ 90\% } & \multicolumn{2}{c}{ 95\% } \\
Accuracy & top1 & top5 & top1 & top5 & top1 & top5 & top1 & top5 \\ \midrule
Unpruned & 73.1 & 91.3 & 73.1 & 91.3 & 75.6 & 92.8 & 75.6 & 92.8 \\ \midrule
SNIP$_{19'}$ & 68.5 & 88.8 & 63.8 & 86.0 & 61.5 & 83.9 & 44.3 & 69.6 \\
GraSP$_{20'}$ & 69.5 & 89.2 & 67.0 & 87.4 & 65.4 & 86.7 & 46.2 & 66.0 \\
FORCE$_{21'}$ & 70.2 & 89.5 & 65.8 & 86.8 & 64.9 & 86.5 & 59.0 & 82.3 \\
DOP$_{22'}$ & - & - & - & - & 64.1 & - & 48.1 & - \\
ProsPr$_{22'}$ & 70.7 & 89.9 & 66.1 & 87.2 & 65.9 & 86.9 & 59.6 & 82.8 \\ \midrule
Sim.Cmb. & 17.3 & 25.8 & 15.4 & 23.0 & 9.9 & 16.4 & 8.3 & 15.3 \\
\textbf{EPSD} & \textbf{71.2} & \textbf{90.1} & \textbf{67.1} & \textbf{87.6} & \textbf{66.3} & \textbf{87.3} & \textbf{60.1} & \textbf{83.0} \\
\bottomrule
\end{tabular}}
\caption{Comparing test accuracy of various advanced early pruning methods at 90\% and 95\% sparsity on full ImageNet. `Sim.Cmb.' refers to the `Simple Combination'.}
\label{tab:cmp_pr_imagenet} 
\end{table}

\noindent \textbf{CIFAR-10/100}.
We perform extensive comparisons with recent early pruning methods on CIFAR-10 and CIFAR-100, and we applied EPSD to two popular lightweight networks (MobileNet-v2~\cite{DBLP:conf/cvpr/SandlerHZZC18} and MobileViT~\cite{DBLP:conf/iclr/MehtaR22}), which is not a common practice in previous early pruning works. We also investigated the iterative version of EPSD. Please refer to the Appendix.

\noindent \textbf{ImageNet}.
We evaluated EPSD on the challenging full ImageNet dataset.
Table~\ref{tab:cmp_pr_imagenet} compared EPSD with advanced pruning methods in terms of top-1 and top-5 accuracy under $90\%$ and $95\%$ sparsity ratio with VGG-19 and ResNet-50.
EPSD surpasses other early pruning methods and notably addresses the degradation problem of `Simple Combination' on challenging datasets. 
For instance, EPSD leads GraSP by $0.9\%$ and improves by $0.4\%$ over ProsPr at sparsity 90\% with ResNet-50.
This highlights EPSD's effective synergy of early pruning and SD, leading to enhanced performance.

\noindent \textbf{Structured Pruning}.
To illustrate the extensibility of EPSD, We evaluate structured pruning, where entire channels are eliminated rather than individual weights.
We compare EPSD against 3SP~\cite{DBLP:journals/corr/abs-2007-00389}, ProsPr~\cite{alizadeh2022prospect}, and random structure pruning reported in ProsPr.
The results are summarized in Table~\ref{tab:structured_pr_a}, and our EPSD achieves the best accuracy performance compared with other structured pruning methods.

\subsection{Comparison of SD Methods}
\label{sec:4.3}
Since EPSD is to explore the~\textit{effective combination} of early pruning and SD, we compare EPSD with SD methods to show the effectiveness.
Specifically, we compare EPSD with LSR~\cite{szegedy2016rethinking}, TFKD~\cite{yuan2020revisiting}, CSKD~\cite{yun2020regularizing}, PSKD~\cite{kim2021self}, and DLB~\cite{shen2022self} using various models (ResNet-32/110 and VGG-16/19) on CIFAR-10/100.
When compared to SD methods, EPSD prunes networks at 80\% sparsity.
Table~\ref{tab:cifar_cmp_skd} shows the comparison results.
Surprisingly, though EPSD removes most of the weights, it still achieved comparable or better performance than other advanced SD methods.
Please be aware that directly comparing early-pruned models with unpruned self-distilled models is uncommon in prior research.
This is because models obtained through early pruning are often considered less trainable~\cite{lee2019signal,frankle2020pruning,wang2022recent}. 
However, we demonstrate that combining early pruning with self-distillation is a viable and competitive approach.

\begin{table}[ht!]
\centering
\small
\renewcommand\arraystretch{0.8} 
\setlength{\tabcolsep}{2.8mm}{
\begin{tabular}{llcc}
\toprule
Sparsity & Method & CIFAR-10 & CIFAR-100\\ \cmidrule(lr){1-4}
-  & Unpruned & 93.88(\textit{\%}) &  72.84 (\textit{\%}) \\ \cmidrule(lr){1-4}
80\% & Random &  92.00  &  67.50  \\
     & 3SP &  93.40  &  69.90  \\
     & ProsPr &  93.61  &  72.29  \\ \cmidrule(lr){2-4}
     & \textbf{EPSD} & \textbf{93.82} & \textbf{73.16} \\ \cmidrule(lr){1-4}
90\% & Random &  90.40  &  63.80  \\
     & 3SP &  93.10  &  68.30  \\
     & ProsPr &  93.64  &  71.12  \\ \cmidrule(lr){2-4}
     & \textbf{EPSD} & \textbf{93.72} & \textbf{71.80} \\ 
\bottomrule
\end{tabular}}
\caption{Test accuracy among various structured pruning methods using VGG-19 on CIFAR-10 and CIFAR-100 under sparsity ratios 80\% and 90\%.}
\label{tab:structured_pr_a} 
\end{table}

\subsection{Impact of SD-based Pre-training}
\label{sec:sc-2}
In previous sections, we showed that a simple combination of early pruning and SD can lead to performance degradation.
To verify the key idea of EPSD that identifying more distillable weights enhances the accuracy performance, we design another way for combination: 1) start by training the network from scratch with SD, then 2) prune it, and 3) fine-tune the pruned model with SD to regain performance.
We name this method `Simple Combination-2' (SC-2). Compared to `Simple Combination' (SC-1), SC-2 requires more pre-training effort. To explore the potential impact of SD-based pre-training on the pruned model, we tested SC-2's effect on ImageNet using ResNet-50. Experiments shown in Table~\ref{tab:sc-2} indicated that SC-2 achieved \textit{comparable} accuracy to EPSD (66.4\% vs. 66.2\%). We argue this happened because \textit{SD-based pre-training in SC-2 produced distillable weights}. After pruning with a standard cross-entropy (CE) loss, the remaining weights still kept their distillable nature, allowing the pruned model to regain from fine-tuning with SD. In addition, building upon SC-2, we used EPSD to compress the SD-pre-trained model (instead of starting from random initialization), resulting in further accuracy improvement (66.6\%), which is attributed to retaining more distillable weights through pruning with the SD loss.

\begin{table}[!tb]
\centering
\small
\renewcommand\arraystretch{1.0} 
\setlength{\tabcolsep}{1.4mm}{
\begin{tabular}{c|c|cccccc}
\toprule
Net. & U.P. & LSR & TFKD & CSKD & PSKD & DLB & \textbf{EPSD} \\ \midrule
R32 &  93.46  &  93.27  &  93.68  &  93.12  &  94.04  &   94.15  &   \textbf{94.68}   \\
R110 &  94.79  &  94.40  &  95.08  &  93.88  &  94.91  &   95.15  &   \textbf{95.32}  \\
V16 &  93.97  &  94.09  &  94.08  &  93.78  &  94.10  &   \textbf{94.62}  &   94.51  \\
V19 &  93.88  &  93.95  &  94.09  &  93.62  &  93.93  &  94.42 & \textbf{94.45} \\ \midrule
R32 &  71.74  &  71.79  &  73.91  &  70.79  &  72.51  &   74.00  &   \textbf{74.30}   \\
R110 &  76.36  &  76.68  &  72.98  &  76.59  &  77.15  &   78.18  &   \textbf{78.45}  \\
V16 &  73.63  &  74.19  &  74.06  &  74.19  &  74.05  &   76.12  &  \textbf{76.31}  \\
V19 &  74.61  &  73.25  &  72.54  &  73.35  &  73.64  &   75.47  &   \textbf{76.11}  \\
\bottomrule
\end{tabular}}
\caption{Comparing test accuracy against advanced SD methods and the unpruned baseline (U.P.). The top section shows CIFAR-10 and the lower section displays CIFAR-100. We use `R' for ResNet and `V' for VGG. EPSD is 80\% sparsity, while the other approaches remain unpruned.}
\label{tab:cifar_cmp_skd} 
\end{table}

\subsection{Downstream Tasks}
\label{sec:downstream}
We further verify the robustness of EPSD on two downstream tasks presented below. See the Appendix for details.

\noindent\textbf{Weakly Supervised Object Localization}.
\label{subsec:down_wsol}
As shown in Table~\ref{tab:dsm_wsol}, we reported the error rates with a pruning ratio of 50\%. 
Compared to ProsPr, EPSD achieved lower errors (Cls. Err of 24.40\% vs. 25.39\%, Top-1 Loc. Err. as low as 41.23\%). 
Compared to the unpruned baseline, EPSD only saw a slight 0.27\% drop in localization accuracy, showing improved generalization in weakly supervised scenarios.

\noindent\textbf{Semantic Segmentation}.
\label{subsec:down_seg}
As shown in Table~\ref{tab:dsm_seg}, across two different metrics, EPSD outperforms ProsPr and the `Simple Combination'. 
Specifically, EPSD achieves a 1.63\% higher than ProsPr in mean IoU and 2.63\% higher in pixel accuracy. 
Compared to the `Simple Combination', the improvements are even more significant, with increases of 5.26\% and 5.76\% in two metrics, respectively.

\begin{table}
\centering
\small
\renewcommand\arraystretch{0.8} 
\setlength{\tabcolsep}{1.2mm}{
\begin{tabular}{c|c|cc|cc}
\toprule
\multirow{2}{*}{Method} & P.T. w/ SD & \multicolumn{2}{c|}{PR. w/} & \multicolumn{2}{c}{(Re-)Train w/ SD} \\ 
& \#Epochs & CE & SD & \#Epochs & Top1 Acc.(\%) \\ \midrule
SC-1 & 0 & \checkmark & & 100 & 9.9 \\
EPSD & 0 & & \checkmark & 100 & 66.2 \\ \midrule
SC-2 & 100 & \checkmark &  & 100 & 66.4 \\
P.T.+EPSD & 100 &  & \checkmark & 100 & 66.6 \\
\bottomrule
\end{tabular}}
\caption{Investigation of the impact of SD-based Pre-training. `P.T.' means pre-training and `PR.' is the pruning process with a 90\% sparsity ratio. }
\label{tab:sc-2} 
\end{table}

\begin{table}
\centering
\small
\renewcommand\arraystretch{1.3}
\setlength{\tabcolsep}{2.2mm}{
\begin{tabular}{ccccc}
\toprule
  \multirow{2}{*}{Method}   &  \multirow{2}{*}{\textit{s.p.}} & \multirow{2}{*}{Cls.Err. ($\downarrow$)} & \multicolumn{2}{c}{Loc.Err. ($\downarrow$)} \\ 
  & & & Top-1 & Gt-Known \\ \cmidrule(lr){1-5}
  Unpruned &    -        & 23.90\%    & 40.96\%    & 23.97\% \\ \cmidrule(lr){1-5}
  ProsPr   &    50\%     & 25.39\%    & 48.65\%   & 32.69\% \\
  Sim.Cmb. &    50\%     & 27.53\%    & 50.57\%   & 33.33\% \\
  EPSD & 50\%   & 24.40\%    & 41.23\%   & 25.08\% \\ 
\bottomrule
\end{tabular}} 
\caption{Results of weakly supervised object localization task on CUB-200-2011. The top-1 classification error (Cls.Err.)  and localization error rates (Loc.Err.) are reported.}
\label{tab:dsm_wsol} 
\end{table}

\begin{table}
\centering
\small
\renewcommand\arraystretch{1.3}
\setlength{\tabcolsep}{3.2mm}{
\begin{tabular}{cccc}
\toprule
  Method   &  \textit{s.p.} & Mean IoU ($\uparrow$) & pixAcc ($\uparrow$) \\ \cmidrule(lr){1-4}
  Unpruned &    -      & 46.46\%    & 85.70\%   \\ \cmidrule(lr){1-4}
  ProsPr   &    40\%   & 42.87\%    & 80.34\%   \\
  Sim.Cmb. &    40\%   & 39.24\%    & 77.21\%   \\
  EPSD     &    40\%   & 44.50\%    & 82.97\% \\
\bottomrule
\end{tabular}}
\caption{Results of semantic segmentation task on Pascal VOC 2012. The mean intersection-over-union (Mean IOU) and pixel accuracy (pixAcc) are reported. }
\label{tab:dsm_seg} 
\end{table}

\subsection{Discussion and Limitation}
\label{sec:takeaway}
This paper explores an efficient model compression framework. By effectively combining early pruning with SD, EPSD improved performance for pruned models without the burden of extensive training.
Importantly, we address the degradation issue arising in a simple combination of early pruning and SD, shedding light on a promising research direction for combining these two techniques, which might offer enlightening insights to the community.
However, this paper mainly addresses fundamental vision models in computer vision. Our focus has yet to encompass the presently prevalent large-scale language or multi-model networks. It remains a potential direction for our future research.

\section{Conclusion}
In this study, we introduce the Early Pruning with Self-Distillation (EPSD) framework, which identifies and retains distillable weights during pruning for a specific SD task. EPSD seamlessly integrates early pruning and SD in just two steps, ensuring the trainability of pruned networks for effective model compression.
We unveil that a straightforward combination of pruning and SD can result in performance decline, particularly at high sparsity ratios. 
Extensive visual and quantitative analysis show that EPSD enhances the trainability of pruned networks, and outperforms advanced pruning and SD methods.
We believe EPSD will inspire more follow-ups for efficient compression of other multi-modal networks, which will accelerate the deployment of the latest deep models to edge devices.



\bibliography{aaai24} 
%

\appendix
\clearpage
{\huge \bf Appendix}

\setcounter{equation}{0}
\setcounter{figure}{0}
\setcounter{table}{0}
\renewcommand{\thetable}{A\arabic{table}}
\renewcommand{\thefigure}{A\arabic{figure}}


We provided more implementation details and additional results for the proposed Early Pruning with Self-Distillation (EPSD) in this appendix.
Algorithm~\ref{alg:psd_alg} showed the detailed compression procedure of our method.

The organization of the appendix is as follows:
\begin{itemize}[itemsep=1pt,topsep=1pt,parsep=1pt,leftmargin=12pt]
\item
    In Sec.~\ref{appendix_sec:setup}, we outlined the datasets, networks, and other experimental setups employed in this paper.
\item 
    Sec.~\ref{sec:supp_empirical_study} provided the experimental setups for the empirical studies shown in the main manuscript, to underscore the credibility of our research.
\item 
    In Sec.~\ref{sec:supp_add_exps}, we present additional experimental results, including the theoretical training/test FLOPs, the combined effects with traditional knowledge distillation (KD), and a brief investigation when integrating SD with a dynamic sparse training (DST) method.
\item 
    Sec.~\ref{sec:supp_cmp_pruning_cifars} presented comparison results of EPSD against other advanced early pruning approaches on CIFAR-10 and CIFAR-100 datasets. Furthermore, we also assessed the performance of EPSD using recently popular lightweight networks.
\item 
    In Sec.~\ref{sec:eff_iter}, we introduced the iterative manner of EPSD, referred to as EPSD-It. We demonstrate the effectiveness of EPSD-It through a comprehensive comparison of various pruning methods (post-training pruning and early pruning) across different levels of sparsity. Moreover, we evaluate EPSD-It in contrast to EPSD, emphasizing its improved confidence estimation capabilities.  
\item 
    In Sec.~\ref{sec:imagenet-sd}, we presented additional results of EPSD on full ImageNet, showcasing the robustness and generalization of our approach to challenging large-scale datasets.
\item  
    In Sec.~\ref{sec:supp_4.1}, we provided the implementation details and more results on EPSD equipped with three SD methods~\cite{yun2020regularizing, kim2021self, shen2022self} as discussed in the main manuscript.
\item 
    Sec.~\ref{sec:appendix_abl} presented the ablation study of EPSD. We unveil the efficacy of EPSD by varying the optimization objectives (standard cross-entropy loss vs. SD loss) in two steps. Moreover, we investigate the impact of employing another early pruning method SNIP~\cite{lee2018snip} on EPSD.

\end{itemize}

We aim for readers to develop a more comprehensive understanding of the proposed method through the provision of additional details and experimental insights.

\begin{algorithm}[!tb]
  \caption{Model Compression Procedure of EPSD}
  \label{alg:psd_alg}
  \begin{algorithmic}[1]
    \Require
        $k,j \gets 0$;
        random initialized network $\theta_{init}$;
        pruning mask $m$, target sparsity $S$;
        image batches $x \in \mathcal{X}$;
        learning rate $\alpha$;
        update steps $i$ during pruning;
        total training epochs $e$ for SD.
    \Ensure
      Sparse, converged network $\hat{\theta}_{cmp}$.
    \State Initial $\theta_{0} \gets m \odot \theta_{init}$. \textcolor{blue}{\Comment{Early Pruning with SD}}
    \While{$k<i$}
        \State Forward to get predictive distribution $\widetilde{P}({x_{k};\theta_{k}})$.
        \State Construct soften targets $\widetilde{P}({\overline{x}_{k}};\overline{\theta}_{k})$ by~\cite{yun2020regularizing,kim2021self,shen2022self}.
        \State Compute SD loss, $\widetilde{\Delta}$ by Eq.(6) and Eq.(7).
        \If{$k=i$}
            \State Compute saliency scores $\widetilde{s}$ by Eq.(8).
            \State Sort and get pruning mask $m$ by $\widetilde{s}$, $S$.
            \State Apply $m$: $\hat{\theta}_{0} \gets \theta_{init} \odot m$; break.
        \EndIf
        \State Update: $\theta_{k+1} \gets \theta_{k}-\alpha \cdot \widetilde{\Delta}$; $k\gets k+1$.
    \EndWhile
    \While{$j<e$} \textcolor{blue}{\Comment{Self-Distillation}}
        \State Forward on full training set $\mathcal{X}$.
        \State Compute $\widetilde{\Delta}$ by Eq.(4), (6) and (7).
        \State Update $\hat{\theta}_{j+1} \gets \hat{\theta}_{j}-\alpha \cdot \widetilde{\Delta}$; $j\gets j+1$.
    \EndWhile
    \State Output compressed model $\hat{\theta}_{cmp}\gets \hat{\theta}_{e}$.
  \end{algorithmic}
\end{algorithm}

\section{Datasets and Networks}
\label{appendix_sec:setup}
We mainly employ three multi-class classification benchmark datasets for comprehensive classification performance evaluations.
The CIFAR-10/CIFAR-100~\cite{krizhevsky2009learning} contains 60,000 RGB natural images of 32$\times$32 pixels from 10/100 classes. 
Each class includes 5,000/500 training samples and 1,000/100 testing samples. We followed the widely-used pre-processing from previous works~\cite{he2016deep, zagoruyko2016wide}.
The Tiny-ImageNet is a subset of ILSVRC-2012, made up of 200 classes. Each class includes 500 training and 50 testing samples, scaled at 64$\times$64. All training images were randomly cropped and resized to 32$\times$32 after the normalization. The test images were only normalized.
ImageNet~\cite{deng2009imagenet} classification dataset comprises 1000 classes. Each class is depicted by thousands of images and we resize them into 256$\times$256 pixels RGB images. The accuracy of ImageNet is computed on the validation set. 
We utilize PyTorch~\cite{paszke2019pytorch} version 1.11.0 within the Python 3.8 environment.

\noindent\textbf{CIFAR-10/100, Tiny-ImageNet:}
The network architectures in the main manuscript are ResNet-18, ResNet-32, ResNet-110~\cite{he2016deep}, VGG-16 and VGG-19~\cite{simonyan2014very}. We use ResNet-18 which modifies the first convolutional layer with kernel size 3$\times$3, strides 1 and padding 1, instead of the kernel size 7$\times$7, strides 2 and padding 3, for image size 32$\times$32 by following~\cite{yun2020regularizing, alizadeh2022prospect}. 
During pruning, we use data from 3 batches, each with 128 samples. This can be done in seconds on a single GPU. In training, we follow a consistent setting of hyper-parameters for the training scheme for a fair comparison, and all CNNs are trained using SGD with a momentum of 0.9, and the learning rate is decayed by a factor of 10. For CIFAR-10, CIFAR-100, and Tiny-ImageNet, we augment training data by applying random cropping (32$\times$32, padding 4), and horizontal flipping following previous setting~\cite{yun2020regularizing,kim2021self,alizadeh2022prospect,shen2022self}.

\noindent\textbf{ImageNet:}
To compare with previous compression methods broadly, we employ ResNet-50~\cite{he2016deep} as the backbone network for fair comparisons and report results obtained by our EPSD at sparsity 90\%. According to~\cite{alizadeh2022prospect,wang2020picking}, the batches must have enough samples from all classes in the dataset. Wang et al.~\cite{wang2020picking} recommend using class-balanced batches sized ten times the number of classes. Because ImageNet is larger and more complex, we require data across several pruning iterations to maintain useful distillable weights for SD. We use 512 batches with 128 samples each, following ProsPr's approximation for gradient computation. This takes only a few minutes on a single GPU, showing high efficiency. After pruning, the model is trained for 100 epochs, starting with a learning rate of 0.1, which decreases by a factor of 10 at the 30th, 60th, and 90th epochs. We resize an image as 256$\times$256 and then perform a random crop to have a 224$\times$224 sized input, augmented with horizontal flipping, color jitter, and lighting. The weight decay was set to 0.0001 and the batch size was 256.

\section{Experimental Setup for Empirical Studies}
\label{sec:supp_empirical_study}
In this section, we thoroughly explain the empirical studies and experimental setups in the main manuscript. The content is arranged in the order of the main manuscript.

\noindent\textbf{Trainability Analysis, Fig. 4 in Sec. 3.2.}
For the visualization of loss surface, we employed open-source visualization tools\footnote{\url{https://github.com/tomgoldstein/loss-landscape}} to intricately illustrate the loss surfaces of pruned models obtained through EPSD and the `Simple Combination', respectively. These visualizations reflect the varying degrees of optimization complexity for the pruned models. The main manuscript showcases the results achieved using ResNet-18 on the CIFAR-100 dataset.
For the `Mean-JSV' curves in Fig. 4, we followed the previous works~\cite{wang2021dynamical, DBLP:conf/iclr/Wang023a} to record and compute the Jacobian singular values during training. The main manuscript showcases the results achieved using ResNet-18 on the CIFAR-100 dataset.

\begin{table*}[tb!]
\centering
\small
\begin{tabular}{llllccccc}
\toprule
    \multirow{2}{*}{Method} & \multirow{2}{*}{Type} & \multirow{2}{*}{Dataset} & \multirow{2}{*}{Model} & \multicolumn{5}{c}{Number of Training Epochs} \\
         &       &              &                      & P.T.(ref.) & P.T.(teacher) & Fine-tune & Distill. & Total \\ \midrule
    EPSD & PR+SD & CIFAR-10/100 & ResNet32/110,VGG16/19   & 0   & 0   & 0    & 200 & 200 \\ \midrule
    DMC  & PR    & CIFAR-10     & VGG16                & 300 & 0   & 150  & 0   & 450 \\
    ReKD & KD    & CIFAR-100    & ResNet110$\rightarrow$ResNet32 & 0   & 240 & 0    & 240 & 480 \\
    PKD  & PR+KD & CIFAR-100    & VGG19                & 0   & 200 & 910  & 200 & 1310 \\
    CPKD & PR+KD & CIFAR-10     & ResNet110            & 0   & 0   & 1400 & 85  & 1485 \\
\bottomrule
\end{tabular}
\caption{
Training epochs used by various compression techniques in Fig. 5. Note that the models and datasets used in EPSD are comparable to those in all other methods.}
\label{appendix_tab:supp_effort_details}
\end{table*}

\noindent\textbf{Training Efforts, Fig. 5 in Sec. 3.3.}
We show the number of training epochs used by various compression methods in Table~\ref{appendix_tab:supp_effort_details}, which are provided by the original papers.
For example, in the case of CPKD~\cite{aghli2021combining}, ResNet-110 and ResNet-164 are utilized for image classification on CIFAR-10. They begin with 6 pruning iterations for the teacher network, fine-tuning for 200 epochs after each iteration to regain initial accuracy. The student network is then trained for 85 epochs using KD, followed by an additional 200-epoch fine-tuning phase. In total, 1485 epochs are employed for comprehensive model compression.
In terms of training time, we replicate the PKD and ReKD using the authors' open-source projects\footnote{\url{https://github.com/dvlab-research/ReviewKD}, \url{https://github.com/ososos888/prune-then-distill}}, maintaining identical hardware environment (using an NVIDIA A40 GPU).

\noindent\textbf{Results on ImageNet, Table 1 in Sec. 4.2.}
In Table 1 of the main manuscript, we report the classification results on full ImageNet with EPSD equipped with SD method DLB. For detailed configurations, please refer to Sec.~\ref{appendix_sec:setup}.

\noindent\textbf{Structured Pruning, Table 2 in Sec. 4.2.}
We further assess EPSD within structured pruning, wherein entire channels are masked. We modified the shape of the pruning mask $m$ to encompass one entry per channel (or column of the weight matrix) as in ProsPr. In Table 2 of the main manuscript, we report the classification results of structured pruning with EPSD equipped with SD method PS-KD.

\noindent\textbf{Comparison with SD Methods, Table 3 in Sec. 4.3.}
We report the results of EPSD equipped with SD method PS-KD for the experiments shown in Table 3. For detailed configurations, please refer to Sec.~\ref{sec:supp_pskd}.

\noindent\textbf{Impact of SD-based Pre-training, Table 4 in Sec. 4.4.}
Following the configurations in Table 1, we further assessed the SC-2 in our experiments.
Specifically, we first train a randomly initialized ResNet-50 on ImageNet for 100 epochs and then prune it to obtain a pruned model. Finally, we retrain the pruned model for 100 epochs using the SD method DLB. The difference between `SC-2' and `P.T.+EPSD' is the loss function (standard cross-entropy or SD loss) used to obtain the saliency scores during pruning.

\noindent\textbf{Weakly Supervised Object Localization.} Weakly Supervised Object Localization (WSOL) is a challenging task in computer vision that involves locating objects in images only by image-level labels. Following previous work~\cite{pan2021unveiling}, we performed experiments using VGG-16 as a backbone network on the CUB-200-2011~\cite{wah2011caltech} dataset. 
We replace the original cross-entropy loss function with the SD loss proposed in DLB and keep other optimization objectives unchanged. We also report the reproduced results with an unpruned backbone network and the experimental results using ProsPr, and the relevant parameters in the pruning process are consistent with our settings on CIFAR-10/CIFAR-100, and detailed parameters can be found in Table~\ref{tab:imp_dlb}.
We report the top-1 classification and localization error, gt-known localization error (considers localization only regardless of classification) on CUB-200-2011~\cite{wah2011caltech} in Table 5. We utilize the open-source code\footnote{\url{https://github.com/Panxjia/SPA_CVPR2021}} for the experiment.

\noindent\textbf{Semantic Segmentation}. Semantic segmentation is a vital computer vision task that categorizes image pixels into different classes based on their semantic meaning.  We prune pre-trained FCN32s~\cite{long2015fully} which employs VGG-16 and then train the pruned network for 50 epochs. We also report the reproduced results with an unpruned model and the results using ProsPr. The detailed settings can be found in Table~\ref{tab:imp_dlb}. We report the intersection-over-union (Mean-IOU) and the pixel accuracy (pixAcc) on Pascal VOC 2012~\cite{everingham2015pascal} in Table 6. We utilize the open-source code\footnote{\url{https://github.com/Tramac/awesome-semantic-segmentation-pytorch}} for the experiment.

\section{Additional Experiments}
\label{sec:supp_add_exps}
\subsection{Theoretical Training/Test FLOPs of EPSD}
\begin{table}[h!]
\vspace{-0.7em}
\centering
\scriptsize
\setlength{\tabcolsep}{1pt}
\renewcommand{\arraystretch}{0.7}
\resizebox{1\columnwidth}{!}
{\begin{tabular}{lcccccccccc}\toprule
Model & Dense & \textit{Sparsity} & SNIP & GraSP & FORCE & DOP & ProsPr & EPSD\scalebox{0.6}{+PS-KD} & EPSD\scalebox{0.6}{+CS-KD} & EPSD\scalebox{0.6}{+DLB} \\ \midrule
\multirow{2}{*}{VGG19} & \multirow{2}{*}{1$\times$(15.1T)} & 90\% & 0.3$\times$ & 0.3$\times$ & 0.3$\times$ & - & 0.3$\times$ & 0.3$\times$ & 0.4$\times$ & 0.4$\times$ \\
 &  & 95\% & 0.15$\times$ & 0.15$\times$ & 0.15$\times$ & - & 0.15$\times$ & 0.15$\times$ & 0.2$\times$ & 0.2$\times$ \\ \midrule
\multirow{2}{*}{ResNet50} & \multirow{2}{*}{1$\times$(3.2T)} & 90\% & 0.3$\times$ & 0.3$\times$ & 0.3$\times$ & 0.3$\times$ & 0.3$\times$ & 0.3$\times$ & 0.4$\times$ & 0.4$\times$ \\
 &  & 95\% & 0.15$\times$ & 0.15$\times$ & 0.15$\times$ & 0.15$\times$ & 0.15$\times$ & 0.15$\times$ & 0.2$\times$ & 0.2$\times$ \\
\bottomrule\end{tabular}}
\caption{Theoretical training FLOPs of EPSD and other baselines in Table 1.}
\vspace{-0.7em}
\end{table}
The baseline methods in Table 1 of the main paper require a fine-tuning process where EPSD does not (EPSD directly uses SD training), thus the training cost of EPSD does not increase significantly.
We report theoretical FLOPs following RigL~\cite{evci2020rigging} which counts FLOPs based on the forward/backward pass in training.
For all baselines in Table 1, the total training FLOPs are $3*f_{s}$, where $f_{s}$ is the FLOPs for a given sparse NN. For EPSD, different SD strategy produces diverse training FLOPs ($3*f_{s}$, $4*f_{s}$, $4*f_{s}$ for PS-KD, CS-KD, and DLB, resp.), while their test FLOPs (without backward pass) remains consistent ($1*f_{s}$).
EPSD equipped with PS-KD presents similar FLOPs as pruning methods.

\subsection{Distill Sparse NN with KD}
\begin{table}[h!]
\vspace{-0.6em}
\centering\scriptsize
\setlength{\tabcolsep}{6pt}
\renewcommand{\arraystretch}{0.8}
\resizebox{1\columnwidth}{!}{\begin{tabular}{ccccc}\toprule
Method &  Teacher Model & Student Model & \textit{Sparsity} & Acc. after KD/SD \\ \midrule
\multirow{2}{*}{SNIP+KD} & \multirow{4}{*}{VGG19 Acc.=72.92\%} & random initialized & 95\% & 71.65\% \\
& & pre-trained & 95\% & 71.67\% \\
\multirow{2}{*}{ProsPr+KD} &  & random initialized & 95\% & 55.04\% \\
& & pre-trained & 95\% & 72.75\% \\ \midrule
EPSD & - & random initialized & 95\% & \textbf{73.81}\% \\
\bottomrule\end{tabular}}
\caption{The simple combination of early pruning and traditional KD.}
\vspace{-0.7em}
\end{table}
We also studied the model performance when simply combining different early pruning methods (SNIP and ProsPr) when using traditional KD.
We first pre-train a teacher model, then prune it, and finally use KD to distill the teacher model into student models with different initializations.
More specifically, we evaluate two initializations for the student model (VGG19) in KD: 
1) random initialization, and 2) initialization from pre-training.
The results on CIFAR-100 indicate that 
1) The pruned student model benefits from pre-training. 
2) Compared to SNIP, a random initialized model pruned by ProsPr struggles in standard KD (71.65\% vs. 55.04\%). 
3) Without pre-training for teacher and student models, EPSD achieved better results.

\subsection{DST with SD}
\begin{table}[h!]
\vspace{-0.7em}
\centering
\scriptsize
\setlength{\tabcolsep}{8pt}
\renewcommand{\arraystretch}{0.5}
\resizebox{1\columnwidth}{!}{
\begin{tabular}{cccccc}\toprule
Model & Dataset & Dense & \textit{Sparsity} & RigL & RigL+EPSD \\ \midrule
\multirow{2}{*}{WideResNet-22-2} & \multirow{2}{*}{CIFAR-10} & \multirow{2}{*}{94.36\%} & 90\% & 92.16\% & \textbf{92.49\%} \\
&  &  & 95\% & 90.14\% & \textbf{90.32\%} \\
\bottomrule
\end{tabular}}
\caption{Performance comparison of RigL and RigL+EPSD under different sparsity ratios.}
\vspace{-1.5em}
\end{table}
We specifically investigated the compatibility of DST with SD.
Specifically, we applied the idea of EPSD to RigL, and the DLB is integrated into the DST process. 
We followed the default settings from \url{https://github.com/nollied/rigl-torch} for RigL while keeping other settings the same as EPSD.
Benefiting from dynamic topology during the training, RigL achieves acceptable performance compared to the dense model at high sparsity (95\%), while integrating the idea of EPSD into RigL can further improve its performance.

\section{Comparison of Early Pruning Methods on CIFAR-10/100}
\label{sec:supp_cmp_pruning_cifars}
\begin{figure}[ht!]
\centering
\begin{subfigure}{0.46\textwidth}
\centering
    \includegraphics[width=.9\columnwidth]{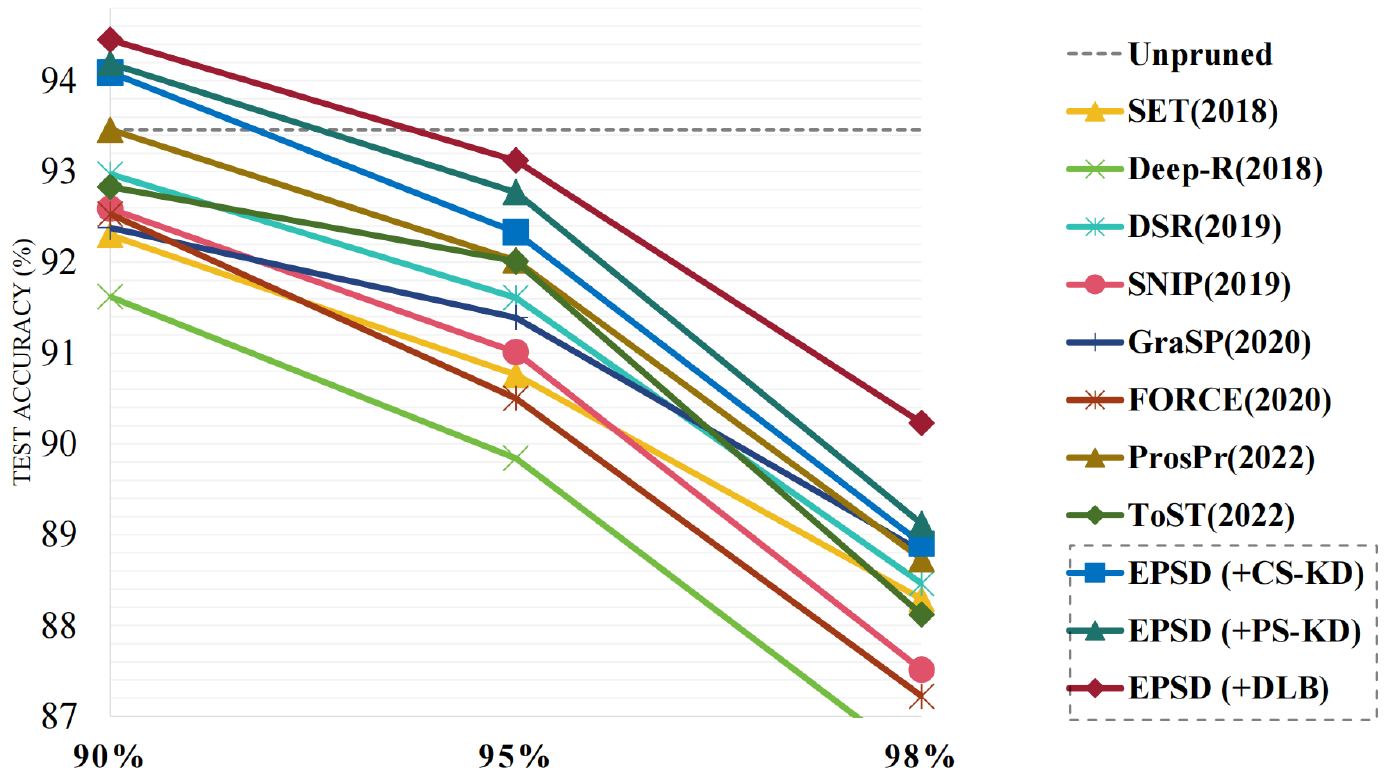}
    \caption{Test accuracy using ResNet-32 on CIFAR-10}
    \label{subfig:cmp_pr_c10}
\end{subfigure} \\
\vspace{6pt} 
\begin{subfigure}{0.46\textwidth}
\centering
    \includegraphics[width=.9\columnwidth]{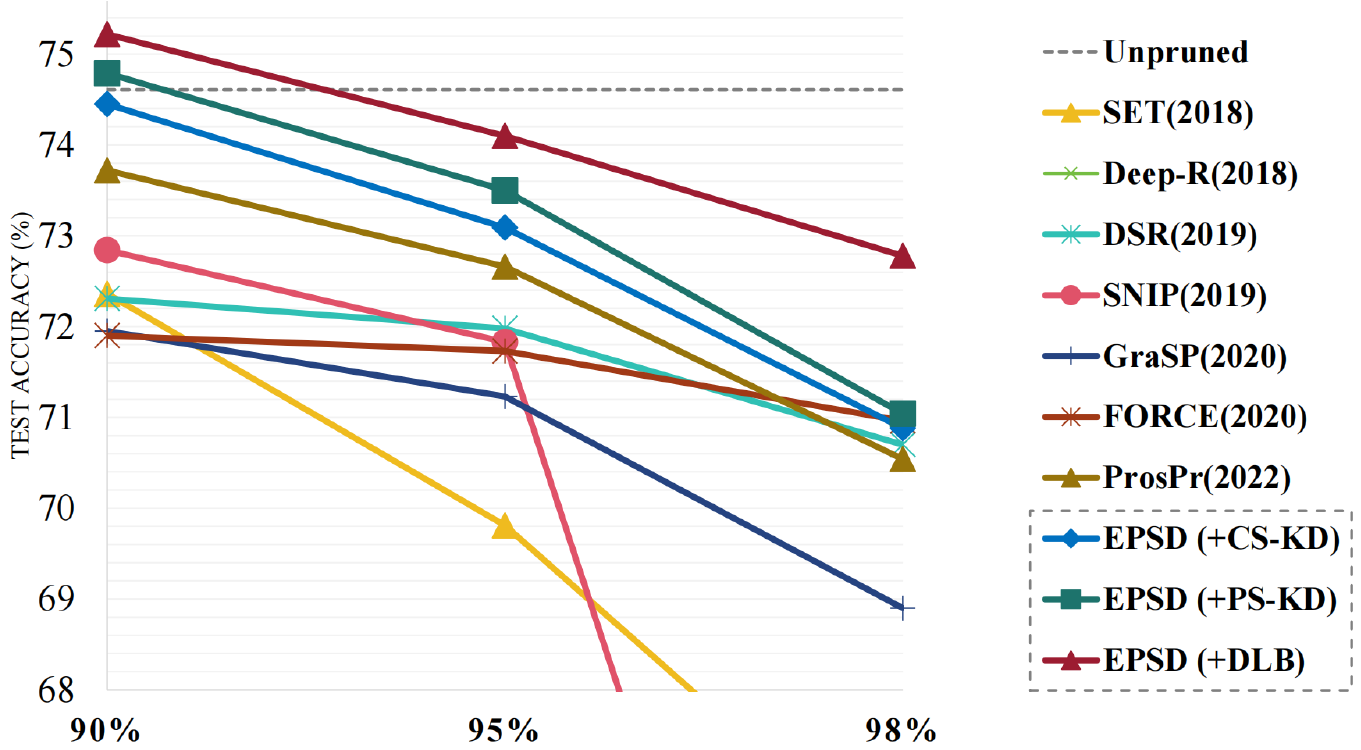}
    \caption{Test accuracy using VGG-19 on CIFAR-100.}
    \label{subfig:cmp_pr_c100}
\end{subfigure}
\caption{
Test accuracy across various advanced pruning methods at sparsity ratios of 90\%, 95\%, and 98\% on CIFAR-10 and CIFAR-100, respectively. Remarkably, EPSD, combined with three integrated SD methods, exhibited superior performance across most cases.
}
\label{fig:cmp_pr_c10_c100}
\end{figure}
We compare EPSD with the early pruning methods 
SNIP~\cite{lee2018snip}, GraSP~\cite{wang2020picking}, FORCE~\cite{de2020progressive}, ToST~\cite{jaiswal2022training} and ProsPr~\cite{alizadeh2022prospect}, and dynamic sparse training methods SET~\cite{mocanu2018scalable}, Deep-R~\cite{bellec2017deep} and DSR~\cite{mostafa2019parameter} on CIFAR-10/100 datasets under sparsity ratios 90\%, 95\%, 98\%, and the accuracy comparison are shown in Fig.~\ref{subfig:cmp_pr_c10} and Fig.~\ref{subfig:cmp_pr_c100}, respectively. The results in Fig.~\ref{fig:cmp_pr_c10_c100} illustrate that EPSD achieved surprising performance at almost all sparsity ratios compared to the advanced early pruning methods. In the case of a 90\% sparsity ratio, EPSD remarkably outperforms the unpruned baselines (94.19\% vs. 93.46\% on ResNet-32 of CIFAR-10 and 74.79\% vs. 74.61\% on VGG-19 of CIFAR-100). Compared with ProsPr, which is most related to our method, EPSD consistently improved the accuracy under various settings. Moreover, results of CIFAR-10/100 showed that the accuracy of EPSD surpassed the unpruned baselines at high sparsity ratios. Overall, EPSD achieves significant performance gains from SD compared to advanced pruning methods.

\begin{table}[!tb]
\centering
\small
\renewcommand\arraystretch{1.2}
\setlength{\tabcolsep}{2.5mm}{
\begin{tabular}{ccccc}
\toprule
    \multirow{2}{*}{Method} & \multirow{2}{*}{Model} & \multirow{2}{*}{U.P.} & \multicolumn{2}{c}{Acc. (\%)} \\ 
            &       &      & 36\% & 59\%  \\ \midrule
     EPSD & \multirow{2}{*}{MobileNet-v2} & \multirow{2}{*}{56.45} & 58.64 & 48.13 \\
    Sim.Cmb. &  &  & 56.32 & 45.70 \\ \midrule
    EPSD & \multirow{2}{*}{MobileViT} & \multirow{2}{*}{65.56} & 65.17 & 64.61 \\
    Sim.Cmb. & & & 65.09 & 63.16 \\
\bottomrule
\end{tabular}
}
\caption{
Test accuracy (\%) of EPSD and the `Simple Combination' with MobileNet-v2 and MobileViT on CIFAR-100, respectively. `U.P.' represents the unpruned baseline.
}
\label{tab:cmp_pr_newbb}
\end{table}

Additionally, we present the results of EPSD employing recently prominent lightweight models (MobileNet-v2 and MobileViT) across varying sparsity in Table~\ref{tab:cmp_pr_newbb}.
For MobileNet-v2\footnote{https://github.com/tonylins/pytorch-mobilenet-v2}, we engage in both pruning and training from scratch, utilizing identical training settings as outlined in Table~\ref{tab:imp_pskd}. For MobileViT, we utilized the model following timm~\cite{rw2019timm}.
The results show that using newer lightweight models, EPSD maintains higher classification accuracy than the `Simple Combination' approach at two sparsity ratios. 
For instance, with MobileNet-v2, EPSD consistently leads the `Simple Combination' by an average of 2.38\% across two sparsity ratios. Notably, EPSD achieves even better results than the unpruned baseline (58.64\% vs. 56.45\%) at a sparsity of 36\%.

\section{Iterative Version of EPSD}
\label{sec:eff_iter}
EPSD can also perform pruning-SD cycles iteratively. Given a target sparsity, each pruning removes some of the weights, then uses SD to restore network performance, and repeats the process until the target sparsity is reached. In the experiment, we combine PS-KD as a study case to explore the performance of the iterative manner of EPSD.

\begin{figure}[ht!]
    \centering
    \begin{subfigure}{0.46\textwidth}
    \centering
        \includegraphics[width=.9\columnwidth]{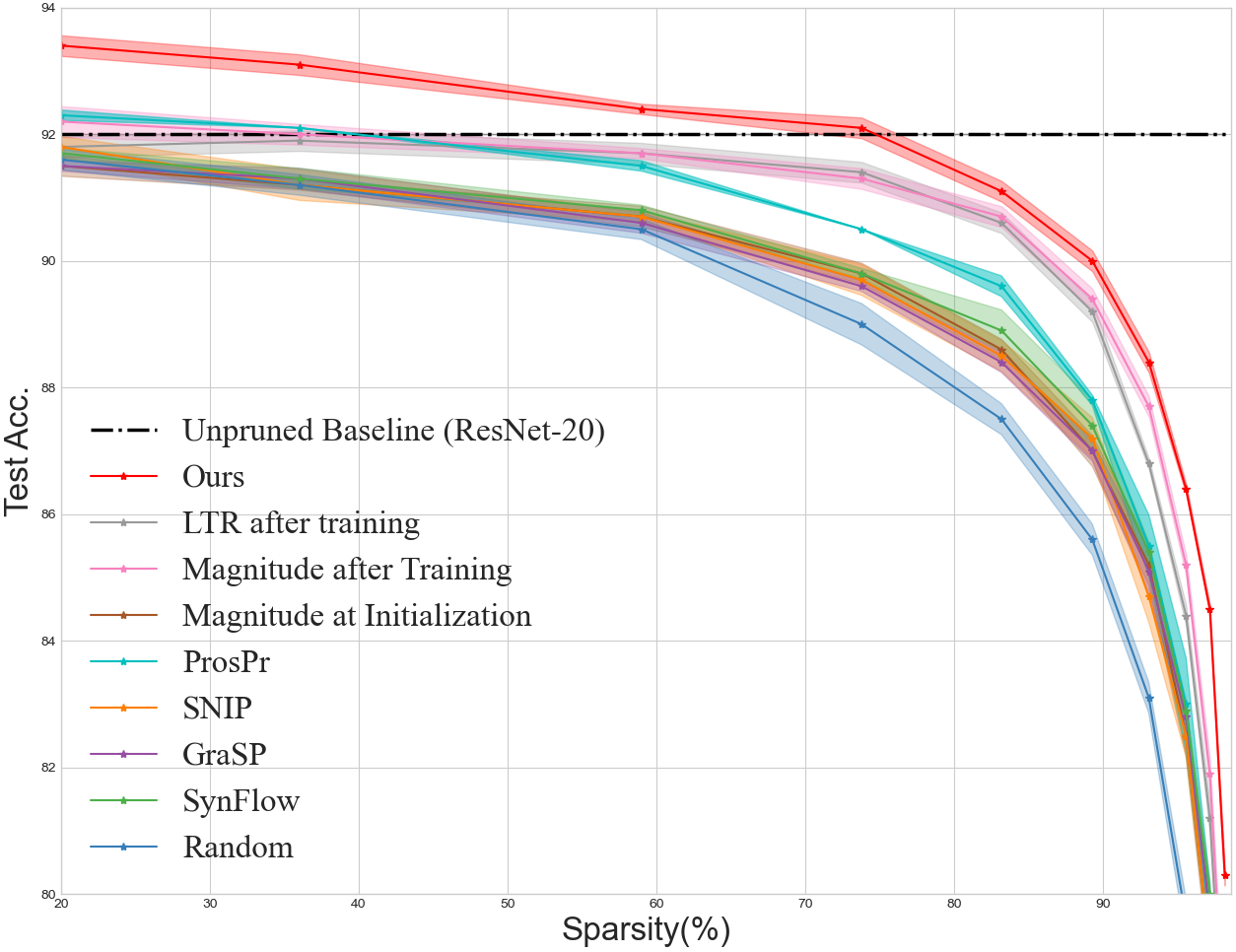}
        \caption{Baseline Network: ResNet-20}
    \end{subfigure} \\
    \vspace{8pt} 
    \begin{subfigure}{0.46\textwidth}
    \centering
        \includegraphics[width=.9\columnwidth]{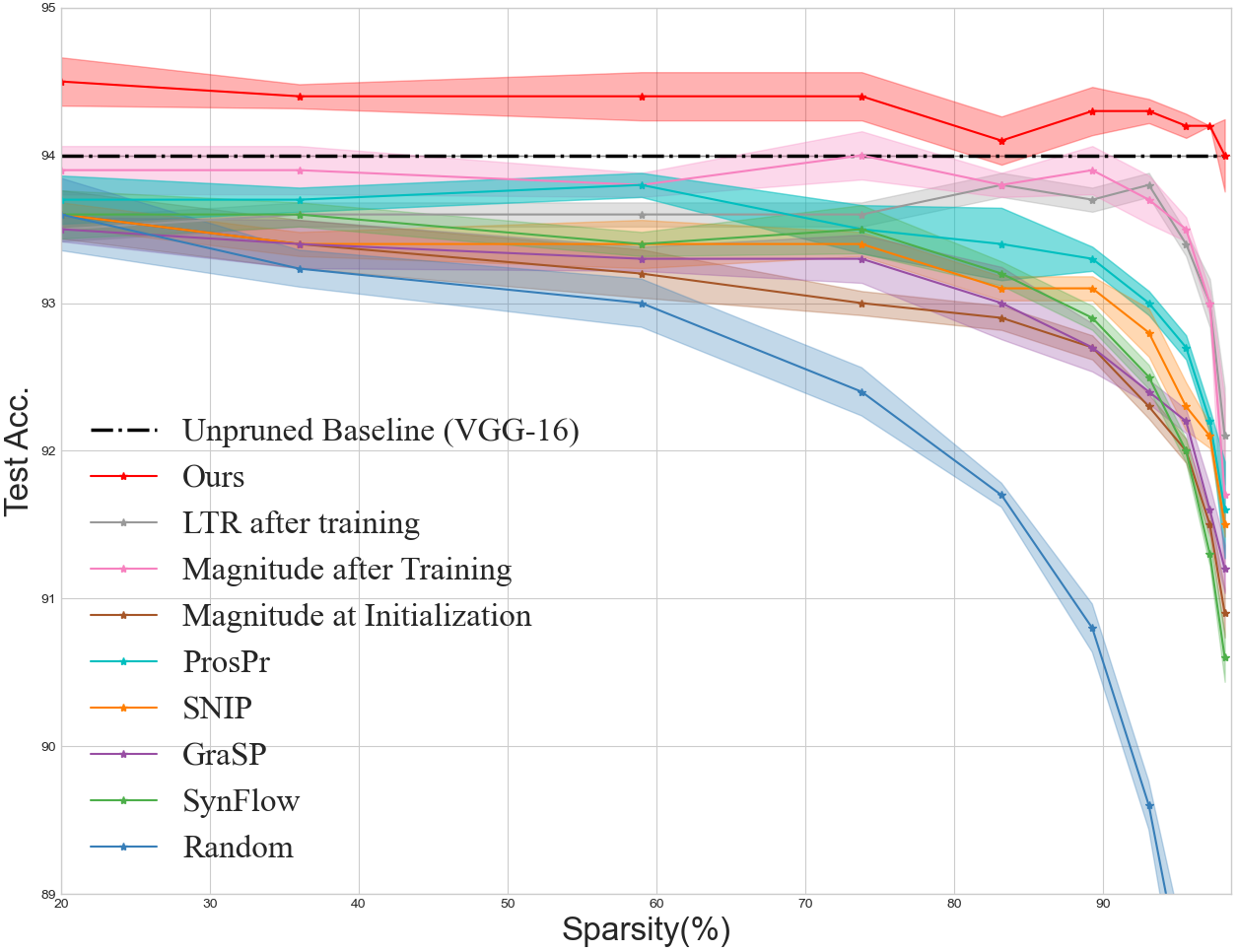}
        \caption{Baseline Network: VGG-16}
    \end{subfigure}
    \caption{
        Test accuracy of EPSD-It against other early pruning and pruning-after-training methods on CIFAR-10 when pruning to various sparsity ratios. 
        The black dashed line represents the unpruned baseline and the shaded areas denote the standard deviation of three runs.
    }
    \label{fig:apdix_cmp_iter}
\end{figure}

\subsection{Comparison of EPSD-It with Advanced Early Pruning Methods.}
Recent work~\cite{frankle2020pruning} assesses the efficacy of early pruning methods under iterative sparsity ratios. They introduce a benchmark that involves training a network by cycling through the complete learning rate schedule anew after every pruning step. To underscore the effectiveness of our iterative pruning approach, EPSD-It achieves accuracy recovery in fewer epochs (30 epochs in our experimentation), as opposed to the entire training timeline, following each pruning iteration. Furthermore, it conducts a full training cycle solely after the final pruning step. We evaluate EPSD-It using ResNet-20 and VGG-16 on CIFAR-10 respectively and show the performance compared to pruning-after-training (PaT) methods (lottery ticket hypothesis after training~\cite{renda2020comparing}, magnitude after training~\cite{frankle2020linear}), early pruning methods (SNIP~\cite{lee2018snip}, GraSP~\cite{wang2020picking}, SynFlow~\cite{tanaka2020pruning}, ProsPr~\cite{alizadeh2022prospect}), and random pruning method in Fig.~\ref{fig:apdix_cmp_iter}. 
It can be found that EPSD-It outperforms other pruning methods, including PaT and early pruning methods, which are trained with a full training schedule after pruning. It is worth mentioning that previous early pruning methods have not yet surpassed PaT pruning methods in most settings. ProsPr is the first attempt to bridge the gap with PaT methods, while our EPSD-It surpasses PaT methods at most sparsity ratios, showing a significant improvement.

\subsection{EPSD vs. EPSD-It}
\begin{figure*}[!t]
\centering
\includegraphics[width=1.0\linewidth]{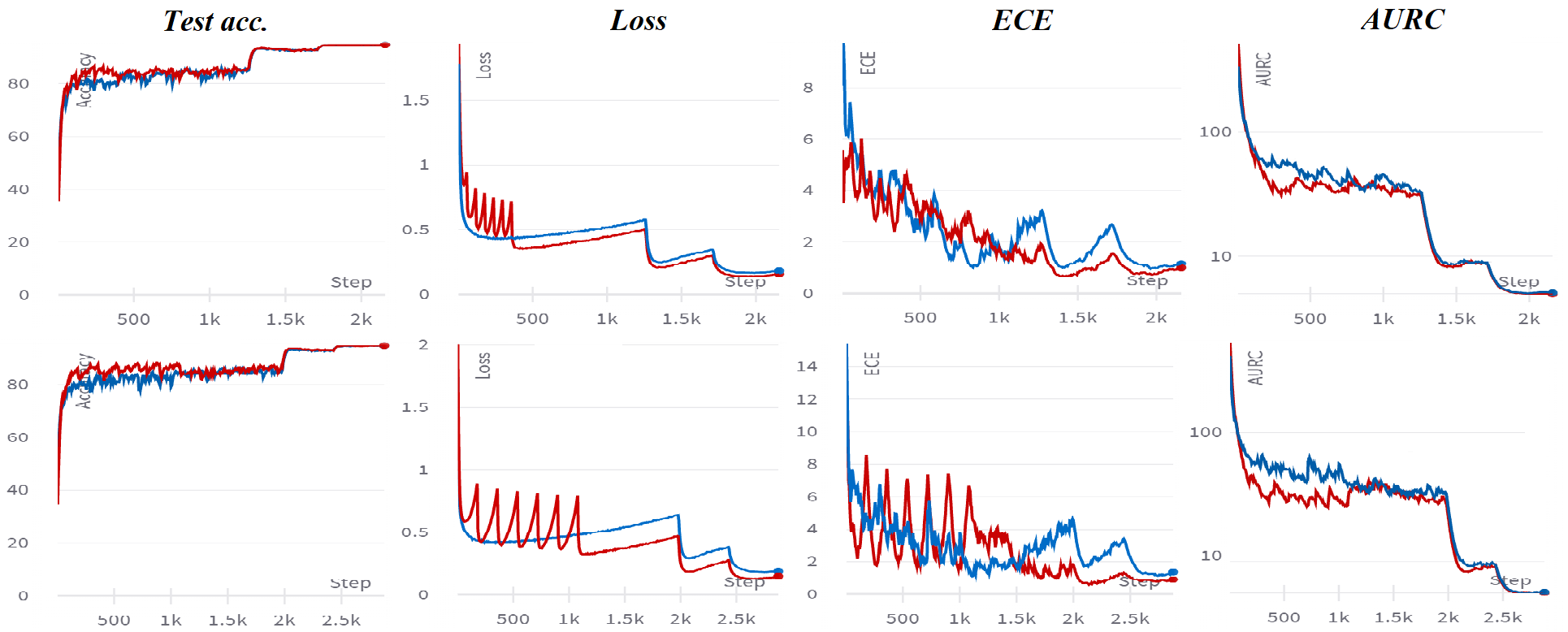}
\caption{Comparison of EPSD ({\color{blue}{blue line}}) and EPSD-It ({\color{red}{red line}}) concerning different metrics on CIFAR-10. The first and second rows correspond to total training epochs of $360$ and $480$, respectively. The horizontal axis represents the training steps and the vertical axes in each row are top-1 test accuracy, training loss, ECE, and AURC, respectively.}
\label{fig:ss_iter}
\end{figure*}
To investigate the performance gain from iteratively performing EPSD, we compare `one-shot EPSD' (EPSD) with the `iterative EPSD' (EPSD-It). Two metrics are adopted to investigate the performance: expected calibration error (ECE)~\cite{naeini2015obtaining} and the area under the risk-coverage curve (AURC)~\cite{geifman2018bias}, to evaluate the quality of predictive probabilities in terms of confidence estimation, following~\cite{kim2021self,yun2020regularizing}. ECE determines whether predictions are well-calibrated, approximating the difference in expectation between classification accuracy and confidence estimates. AURC measures the area under the curve from plotting the risk (i.e., error rate) according to coverage and lower AURC implies that correct and incorrect predictions can be well-separable based on confidence estimates. The maximum class probability is used as a confidence estimator.

We investigate the performance of EPSD and EPSD-It by evaluating various metrics. We experiment with ResNet-18 on CIFAR-10, and the sparsity ratio is set to 80\% for both EPSD and EPSD-It. EPSD-It iteratively prunes 20\% of the remained weights over seven times to reach the target sparsity. In the first 6 prunings, the performance is recovered with a fixed 10 or 30 training epochs (corresponding to the two rows of Fig.~\ref{fig:ss_iter}) and finally train 300 epochs to fully regain performance for EPSD-It. Under this setting, we trained a total of 360 and 480 epochs and kept the consistency of the learning rate decay points in both methods. We observed that EPSD-It gains the benefits of iterative pruning and achieves better results than EPSD during training: 
1) EPSD-It converged the loss faster and achieved lower training errors than EPSD. 2) Regarding the ECE and AURC, EPSD-It eventually converged to a smaller value than EPSD, indicating that EPSD-It has an advantage in confidence estimation.

\section{More Results on Full ImageNet}
\label{sec:imagenet-sd}
\begin{table}
\centering
\small
\renewcommand\arraystretch{1.2}
\setlength{\tabcolsep}{1.6mm}{
    \begin{tabular}{ccccc}
    \toprule
        Acc.(\%) & Pruning Only & EPSD$_{cskd}$ & EPSD$_{pskd}$ & EPSD$_{dlb}$ \\ \midrule
        Top-1 & 65.9 & 64.8 & 66.2 & \textbf{66.3} \\
        Top-5 & 86.9 & 86.6 & 87.2 & \textbf{87.3} \\
    \bottomrule
    \end{tabular}
}
\caption{Performance of EPSD equipped with different SD methods (CS-KD, PS-KD, and DLB) on ImageNet validation set. We utilize ResNet-50 as the baseline network and report the results with a $90\%$ sparsity ratio.}
\label{tab:cmp_sd_imagenet}
\end{table}
Besides the success of EPSD on CIFAR-10/100 and Tiny-ImageNet, we further extend the reference experiment on the large-scale dataset ImageNet~\cite{deng2009imagenet}.
Table~\ref{tab:cmp_sd_imagenet} shows the comparison results for EPSD equipped with three different SD methods with 90\% sparsity. For a fair comparison, the hyperparameters are kept consistent with previous experiments (e.g., $\tau$ and $\alpha$, see Sec.~\ref{appendix_sec:setup} for details). Since DLB~\cite{shen2022self} do not provide results on ImageNet in their papers, we evaluate EPSD to follow consistent basic settings with the other two SD methods and report the Top-1 and Top-5 test accuracy. Therefore, the results reported on ImageNet do not imply that this is the best performance achievable with DLB-equipped EPSD. 
From Table~\ref{tab:cmp_sd_imagenet}, observing that EPSD with DLB (EPSD$_{dlb}$) achieves the best classification accuracy, while EPSD with PS-KD (EPSD$_{pskd}$) has about 0.07\% lower in Top-1 accuracy and EPSD with CS-KD (EPSD$_{cskd}$) achieve 64.75\% Top-1 accuracy rate. The results show that the equipped SD method affects the classification accuracy, while EPSD seems to achieve comparable accuracy on ImageNet when equipped with PS-KD and DLB, respectively. Furthermore, when equipped with PS-KD and DLB, EPSD outperforms the pure pruning method (`Pruning Only') under the same training settings, suggesting that EPSD can also improve accuracy performance on the large-scale dataset.

\begin{figure*}[!ht]
\centering
\begin{tabular}{ccc}
    \includegraphics[width=0.32\textwidth]{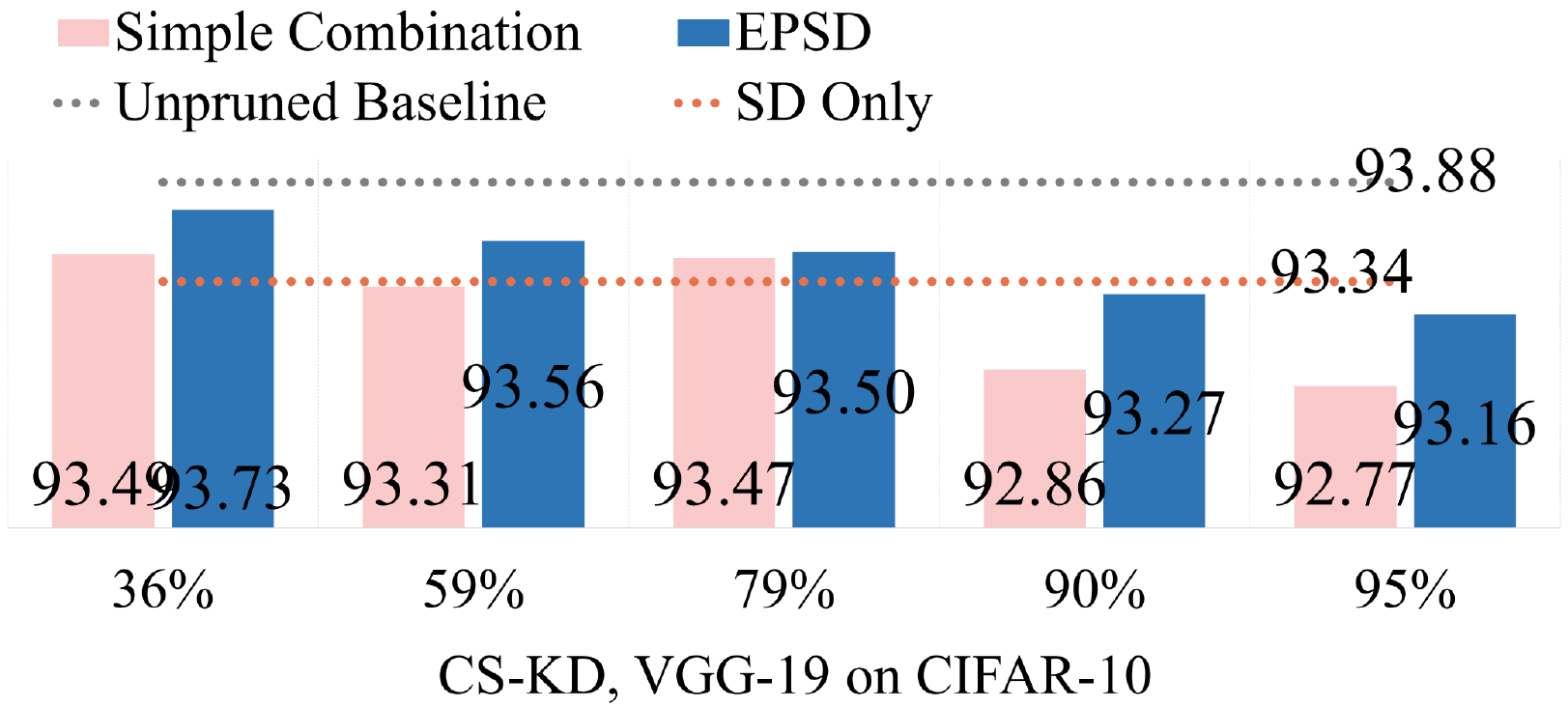} \hspace{-4.2mm} 
    & \includegraphics[width=0.32\textwidth]{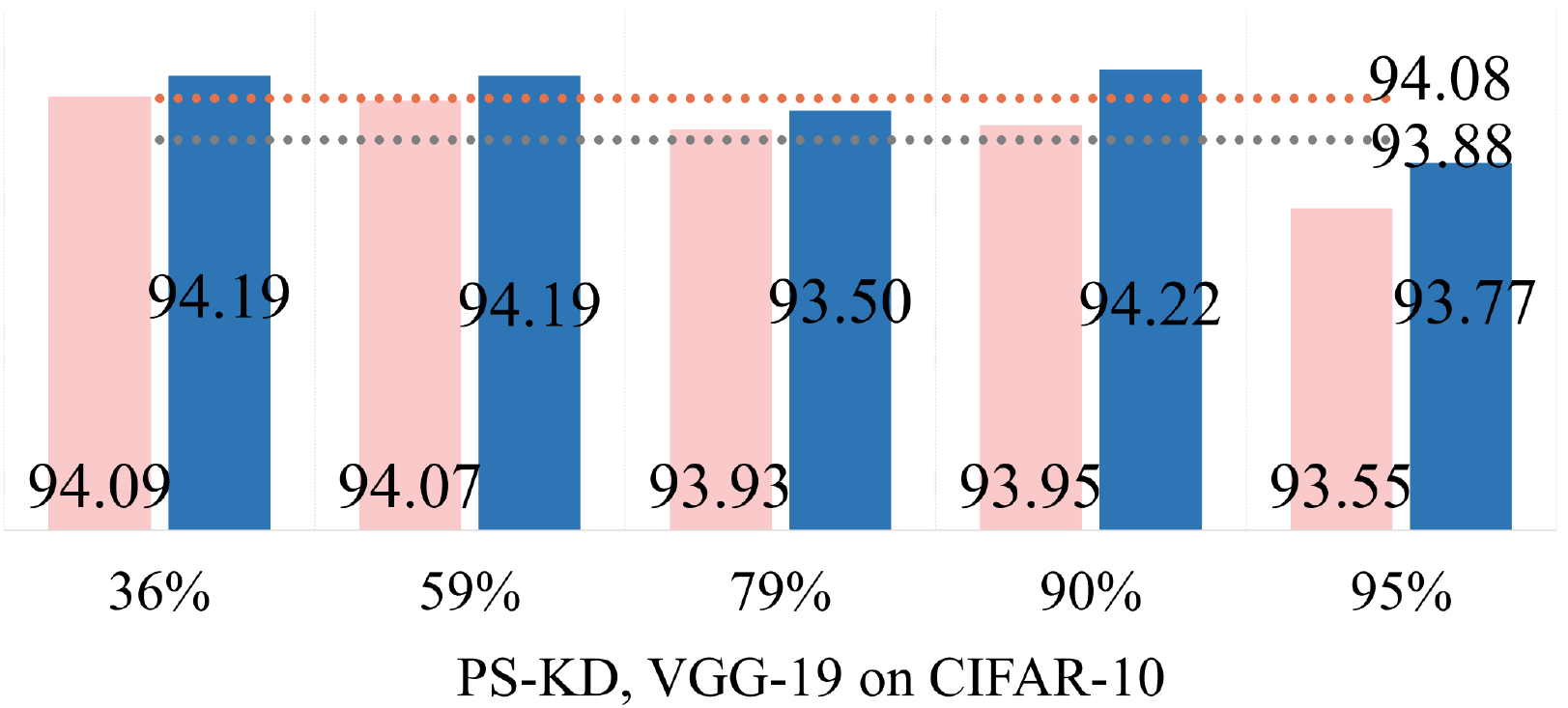} \hspace{-4.2mm}
    & \includegraphics[width=0.32\textwidth]{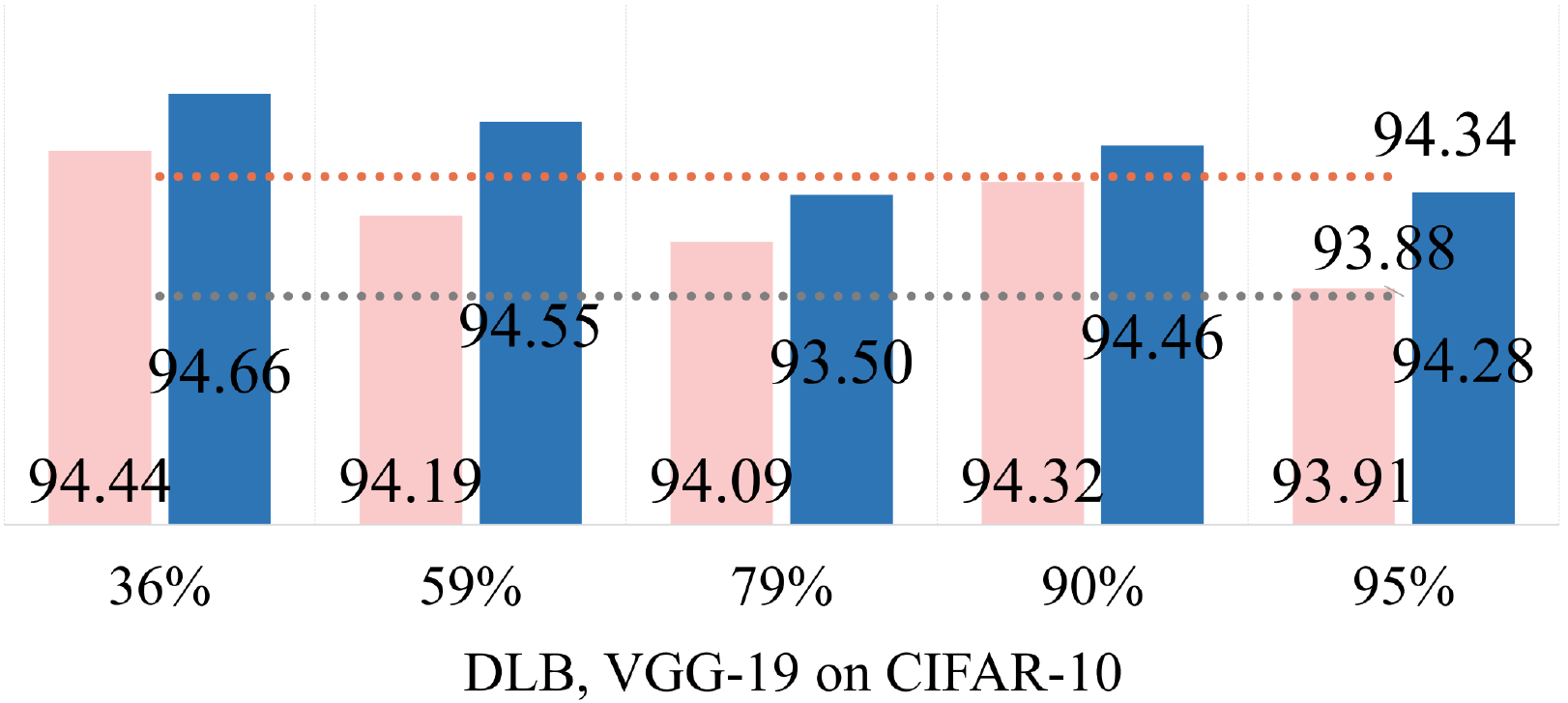} \\
    
    \includegraphics[width=0.32\textwidth]{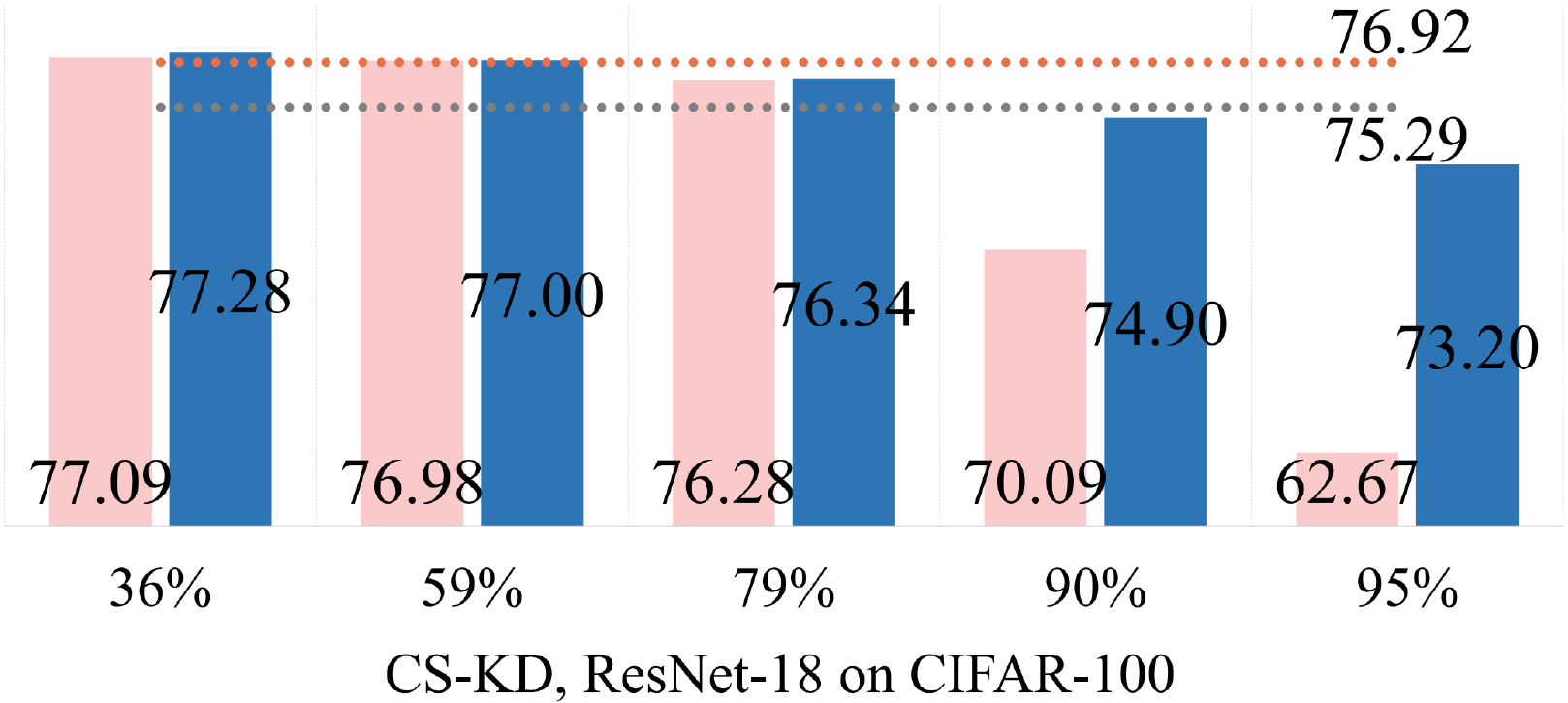} \hspace{-4.2mm}
    & \includegraphics[width=0.32\textwidth]{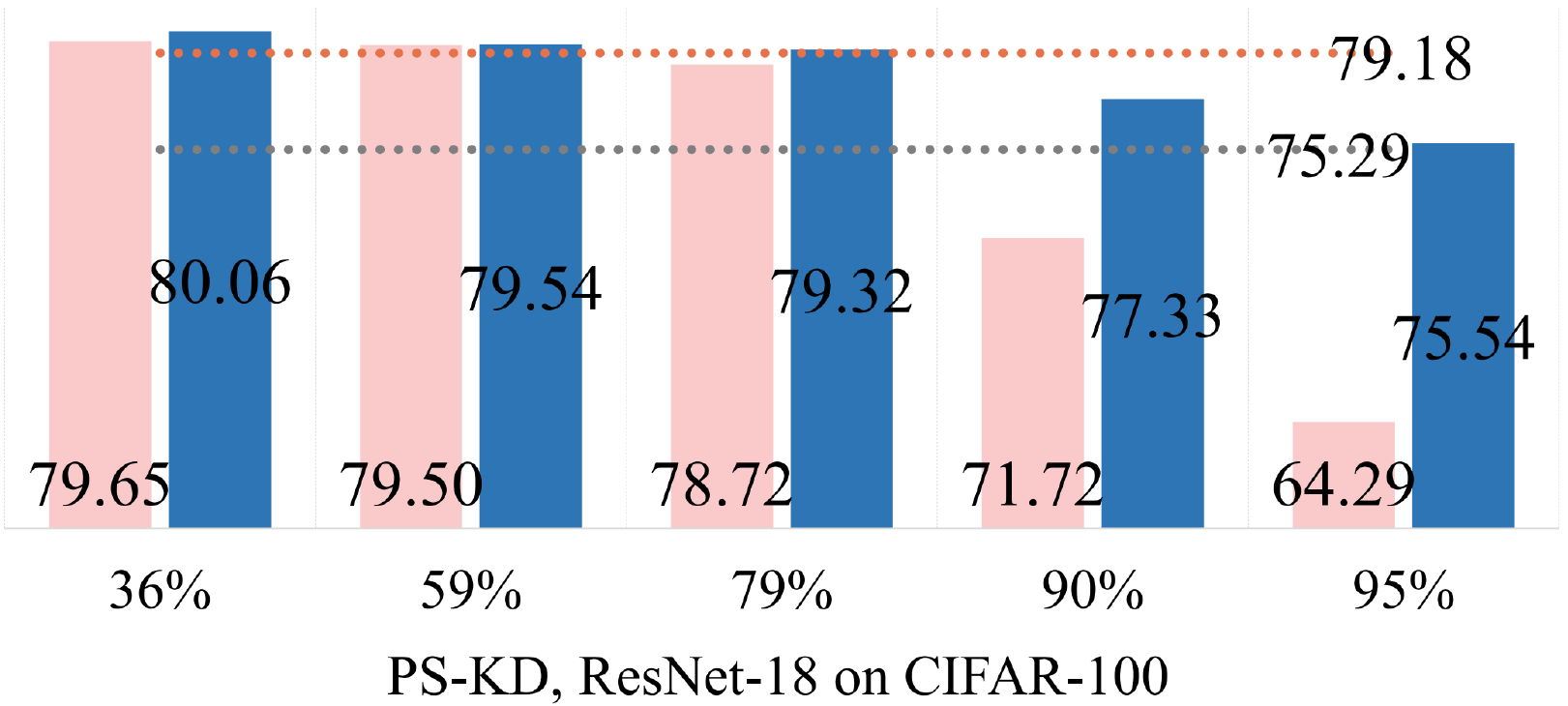} \hspace{-4.2mm}
    & \includegraphics[width=0.32\textwidth]{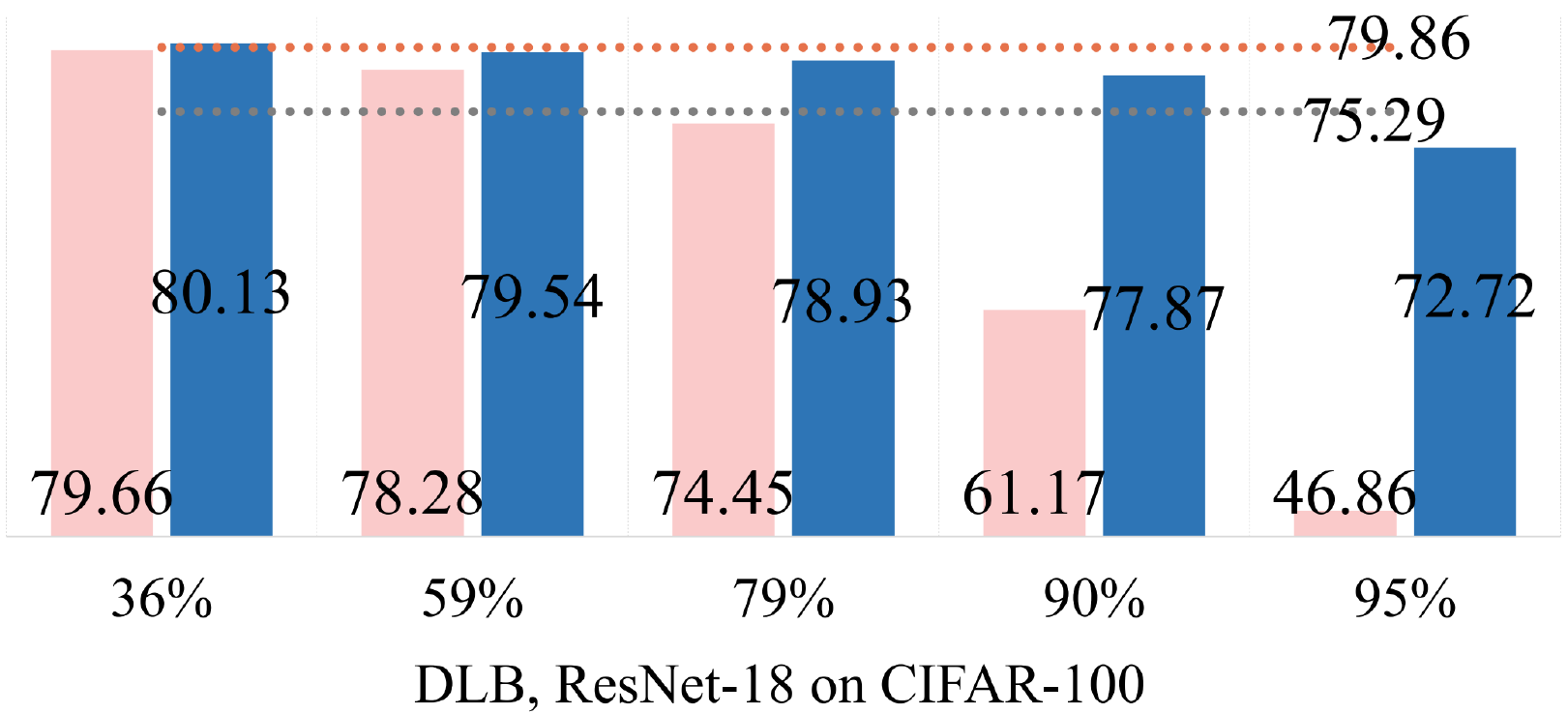} \\
    
    \includegraphics[width=0.32\textwidth]{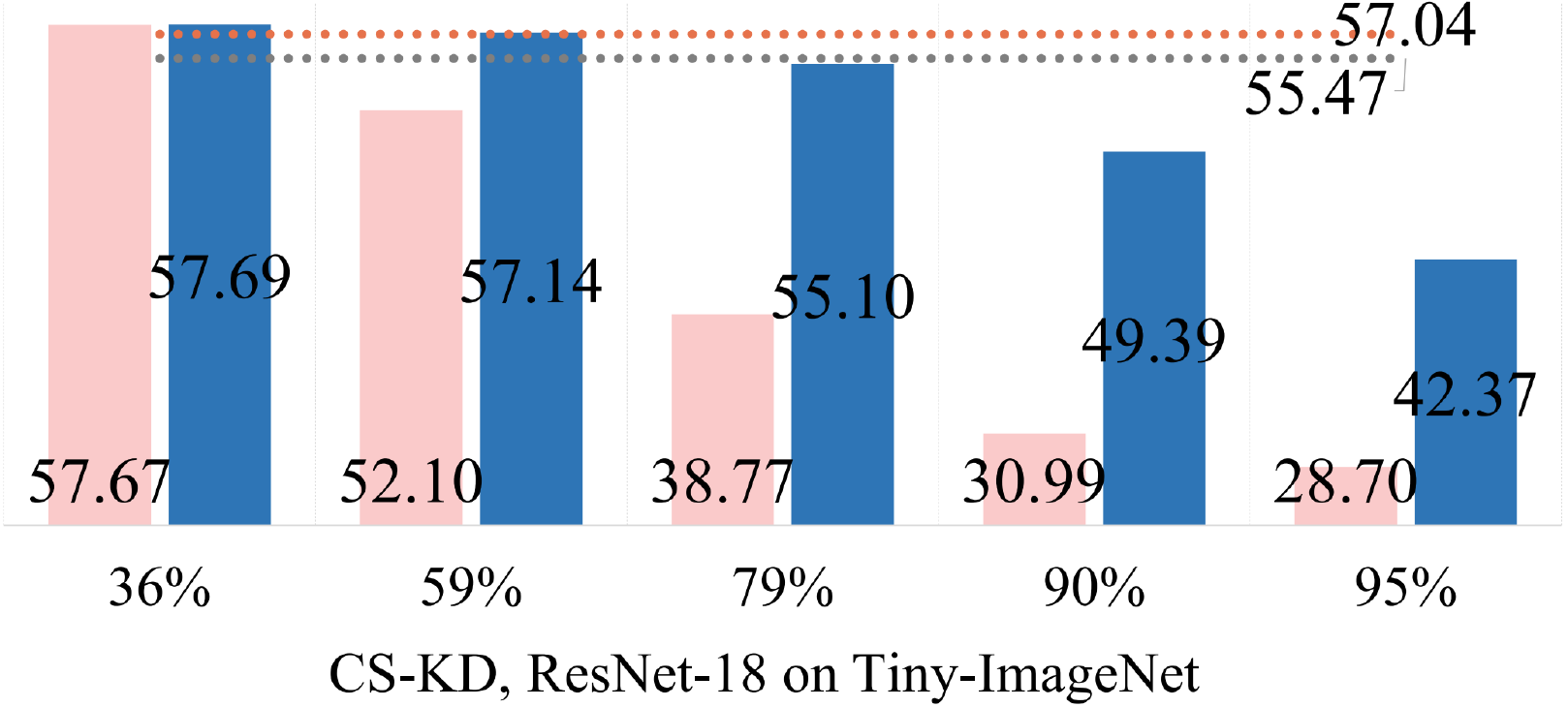} \hspace{-4.2mm}
    & \includegraphics[width=0.32\textwidth]{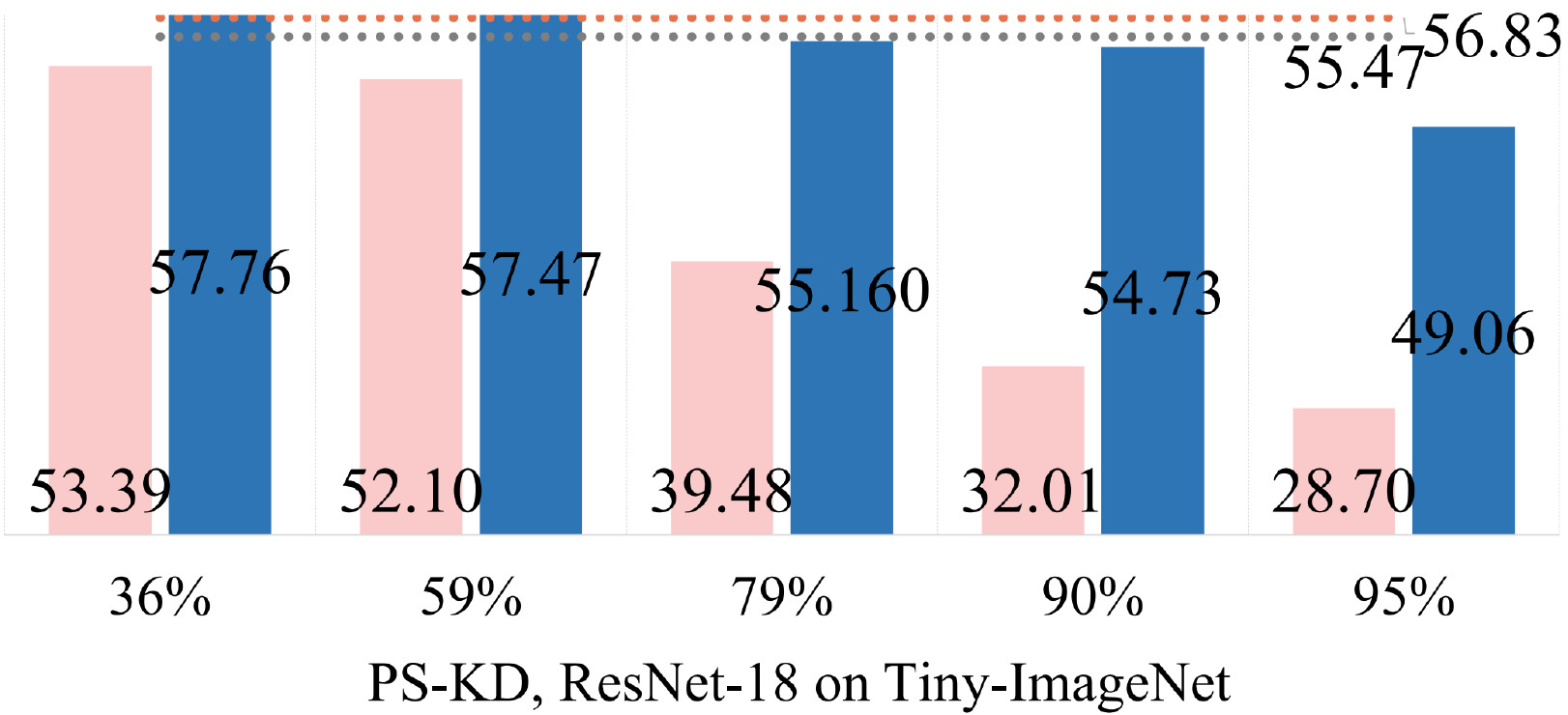} \hspace{-4.2mm}
    & \includegraphics[width=0.32\textwidth]{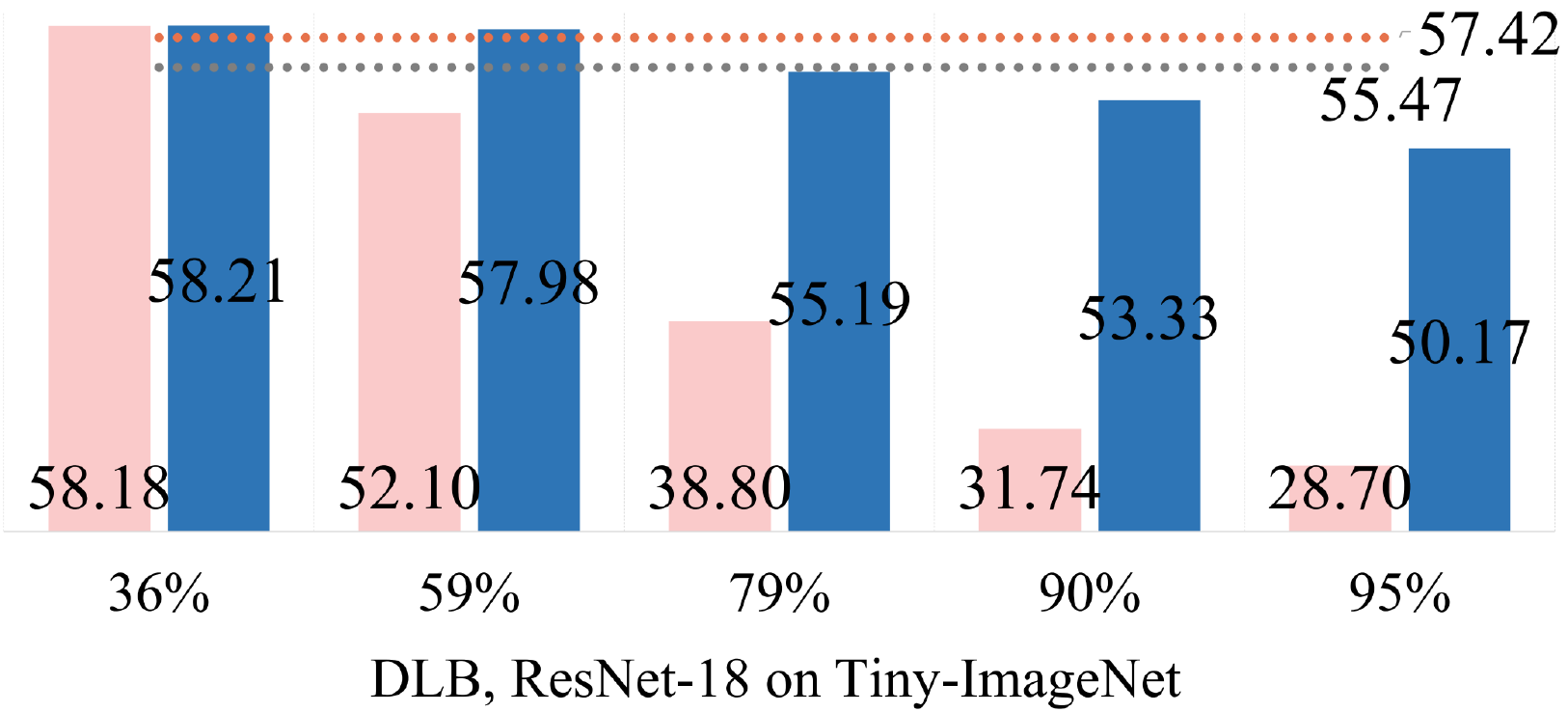} \\
\end{tabular}
\caption{More comparison among the `Simple Combination', `Unpruned Baseline', `SD Only' and EPSD with ResNet-18 and VGG-19 on various datasets (CIFAR-10/100, Tiny-ImageNet) with three equipped SD methods (CS-KD~\cite{yun2020regularizing}, PS-KD~\cite{kim2021self}, DLB~\cite{shen2022self}) under sparsity levels 36\%, 59\%, 79\%, 90\%, and 95\%. 
}
\label{fig:supp_main_results}
\end{figure*}

\begin{table*}[tb!]
\centering
\small
\renewcommand\arraystretch{1.25}
\begin{tabular}{c||ccc}
\toprule
    \textbf{Pruning} & CIFAR-10 & CIFAR-100 & Tiny-ImageNet \\ \hline
    Pruner & ProsPr & ProsPr & ProsPr \\
    Optimizer & nesterov SGD (0.9) & nesterov SGD (0.9) & nesterov SGD (0.9) \\
    Iteration steps & 3 & 3 & 3 \\
    New batch for iteration & \textcolor[RGB]{0,210,0}{\checkmark} & \textcolor[RGB]{0,210,0}{\checkmark} & \textcolor[RGB]{0,210,0}{\checkmark} \\
    Batch size (Iteration) & 128 & 128 & 128 \\
    Learning rate (Iteration) & 0.1 & 0.1 & 0.1 \\
    $\lambda_{cls}$ & 1 & 1 & 1 \\
    Temperature $\tau$ & 4 & 4 & 4 \\ \hline
    \textbf{Self-Distillation} & CIFAR-10 & CIFAR-100 & Tiny-ImageNet \\ \hline
    SD epochs & 200 & 200 & 200 \\
    SD batch size & 128 & 128 & 128 \\
    SD learning rate & 0.1 & 0.1 & 0.1 \\
    LR drop schedule & [100, 150] & [100, 150] & [100, 150] \\
    Drop factor & 0.1 & 0.1 & 0.1 \\
    Weight decay & 0.0001 & 0.0001 & 0.0001 \\
    $\lambda_{cls}$ & 1 & 1 & 1 \\
    Temperature $\tau$ & 4 & 4 & 4 \\
\bottomrule
\end{tabular}
\caption{Hyperparameters of EPSD equipped with CS-KD.}
\label{tab:imp_cskd}
\end{table*}

\begin{table*}[tb!]
\centering
\small
\renewcommand\arraystretch{1.25}
\begin{tabular}{c||ccc}
\toprule
    \textbf{Pruning} & CIFAR-10 & CIFAR-100 & Tiny-ImageNet \\ \hline
    Pruner & ProsPr & ProsPr & ProsPr \\
    Optimizer & nesterov SGD (0.9) & nesterov SGD (0.9) & nesterov SGD (0.9) \\
    Iteration steps & 3 & 3 & 3 \\
    New batch for iteration & \textcolor[RGB]{210,0,0}{\ding{55}} & \textcolor[RGB]{210,0,0}{\ding{55}} & \textcolor[RGB]{210,0,0}{\ding{55}} \\
    Batch size (iteration) & 128 & 512 & 512 \\
    Learning rate (iteration) & 0.1 & 0.1 & 0.1 \\
    $\alpha$ (fixed) & [0.1,0.2,0.3] & [0.1,0.2,0.3] & [0.1,0.2,0.3] \\ \hline
    \textbf{Self-Distillation} & CIFAR-10 & CIFAR-100 & Tiny-ImageNet \\ \hline
    SD epochs & 300 & 300 & 200 \\
    SD batch size & 128 & 128 & 128 \\
    SD learning rate & 0.1 & 0.1 & 0.1 \\
    LR drop schedule & [150, 225] & [150, 225] & [100, 150] \\
    Drop factor & 0.1 & 0.1 & 0.1 \\
    Weight decay & 0.0005 & 0.0005 & 0.0001 \\
    $\alpha$ (linear growth) & 0.8 & 0.8 & 0.8 \\
\bottomrule
\end{tabular}
\caption{Hyperparameters of EPSD equipped with PS-KD.}
\label{tab:imp_pskd}
\end{table*}

\begin{table*}[tb!]
\centering
\small
\renewcommand\arraystretch{1.25}
\begin{tabular}{c||ccc}
\toprule
    \textbf{Pruning} & CIFAR-10 & CIFAR-100 & Tiny-ImageNet \\ \hline
    Pruner & ProsPr & ProsPr & ProsPr \\
    Optimizer & nesterov SGD (0.9) & nesterov SGD (0.9) & nesterov SGD (0.9) \\
    Iteration steps & 3 & 3 & 3 \\
    New batch for iteration & \textcolor[RGB]{0,210,0}{\checkmark} & \textcolor[RGB]{0,210,0}{\checkmark} & \textcolor[RGB]{0,210,0}{\checkmark} \\
    Batch size (Iteration) & 128 & 128 & 128 \\
    Learning rate (Iteration) & 0.1 & 0.1 & 0.1 \\ 
    $\lambda_{cls}$ & 1 & 1 & 1 \\
    Temperature $\tau$ & 3 & 3 & 3 \\ \hline
    \textbf{Self-Distillation} & CIFAR-10 & CIFAR-100 & Tiny-ImageNet \\ \hline
    SD epochs & 240 & 240 & 200 \\
    SD batch size & 64 & 64 & 128 \\
    SD learning rate & 0.05 & 0.05 & 0.2 \\
    LR drop schedule & [150, 180, 210] & [150, 180, 210] & [100, 150] \\
    Drop factor & 0.1 & 0.1 & 0.1 \\
    Weight decay & 0.0005 & 0.0005 & 0.0001 \\
    $\lambda_{cls}$ & 1 & 1 & 1 \\
    Temperature $\tau$ & 3 & 3 & 3 \\
\bottomrule
\end{tabular}
\caption{Hyperparameters of EPSD equipped with DLB.}
\label{tab:imp_dlb}
\end{table*}

\section{EPSD equipped with Various SD Methods}
\label{sec:supp_4.1}
In this section, we provided the detailed experimental setup and more results of EPSD equipped with various SD methods. Specifically, we described the implementation details for EPSD equipped with three SD methods CS-KD~\cite{yun2020regularizing}, PS-KD~\cite{kim2021self} and DLB~\cite{shen2022self}, respectively. We provide more classification results on various datasets, including CIFAR-10, CIFAR-100, and Tiny-ImageNet.

\subsection{EPSD equipped with CS-KD}
\noindent \textbf{Implementations}. In the main manuscript, we have the following definition for the abstract SD loss function:
\begin{equation}
\label{eq:supp_sd_loss}
\mathcal{L}_{S D}=\frac{1}{n} \sum_{i=1}^n \tau^2 \cdot D_{K L}\left(\widetilde{P}\left(\bar{x}_i ; \bar{\theta}_s\right) \| \widetilde{P}\left(x_i ; \theta_s\right)\right),
\end{equation}
where $\widetilde{P}({\bar{x}_{i};\bar{\theta}_{s}})$ represents the soft targets produced by the student networks in SD. For CS-KD~\cite{yun2020regularizing}, Yun et al. propose a class-wise regularization that enforces consistent predictive distributions in the same class. The total training loss of CS-KD is defined as:
\begin{equation}
\label{eq:cskd}
    \mathcal{L}_{\mathrm{CS}-\mathrm{KD}}=\mathcal{L}_{\mathrm{CE}}(x_{i}; \theta_{s})+\lambda_{cls} \cdot \tau^2 \cdot \mathcal{L}_{SD}\left(x_{i},\bar{x}_{i}; \theta_{s}, \bar{\theta}_{s}; \tau\right),
\end{equation}
where $\bar{x}_{i}$ represents another randomly sampled input which having the same classification label and $\bar{\theta}_{s}$ is a fixed copy of the parameters $\theta_{s}$. The $\mathcal{L}_{\mathrm{CE}}$ is the cross-entropy loss and $\tau$ is the temperature coefficient.  A higher temperature results in a more uniform distribution, leading to a similar regularization effect as label smoothing. In EPSD, we utilized Eq.~(\ref{eq:cskd}) in both the pruning and SD phases. We followed the training and validation settings in CS-KD, and the detailed hyperparameters can be found in Table~\ref{tab:imp_cskd}.

\noindent\textbf{More Results}. We provided more results of EPSD combined with CS-KD on different network structures and datasets to supplement the results in the main manuscript, as shown in Fig.~\ref{fig:supp_main_results} (1st column). It can be found that the EPSD consistently outperformed the ‘Simple Combination’ in all settings. For instance, with ResNet-18 on CIFAR-100 at sparsity 90\%, the accuracy of the `Simple Combination' is 12.62\% lower than the `Unpruned Baseline' (62.67\% vs. 75.29\%), while EPSD achieved 74.90\%, which is comparable to the `SD Only' and the `Unpruned Baseline'.

\subsection{EPSD equipped with PS-KD}
\label{sec:supp_pskd}
\begin{table}[!tb]
\centering
\small
\renewcommand\arraystretch{1.2}
\setlength{\tabcolsep}{1mm}{
    \begin{tabular}{cccccc}
    	\toprule
    	   \textbf{Dataset} & \textbf{Baseline} & $\alpha=0.1$ & $0.2$ & $0.3$ & $0.4$ \\ \midrule
    	   \multirow{2}{*}{CIFAR-10} & \multirow{2}{*}{VGG-16} & \underline{$93.97$} &- &- &- \\
    	   & & $\mathbf{94.44}$ & $94.18$ & $94.15$ & $94.32$ \\ \midrule
    	   \multirow{2}{*}{CIFAR-100} & \multirow{2}{*}{ResNet-18} & \underline{$75.29$} &- &- &- \\ 
    	   & & $79.20$ & $\mathbf{79.44}$ & $78.96$ & $79.14$ \\
    	\bottomrule
    \end{tabular}
}
\caption{Ablation study of the proportion of historical knowledge~($\alpha$) when~\textit{pruning} with PS-KD~\cite{kim2021self}. We report the results of ResNet-18 on CIFAR-100 and VGG-16 on CIFAR-10, respectively, and the sparsity is set to $80\%$. The results of the unpruned baseline network are \underline{underlined} and the best accuracy is in \textbf{bold}.}
\label{tab:abl_cifar_alpha}
\end{table}
\noindent \textbf{Implementations}. In PS-KD~\cite{kim2021self}, the network is trained with the soften targets which are computed as a linear combination of the hard labels (ground-truth) and the past predictions at last \textit{epoch}, which is adjusted adaptively as training proceeds by a hyperparameter $\alpha$. Therefore, the $\bar{x}_{i}$ in PS-KD is the same as $x_{i}$, but $\bar{\theta}_{s}$ refers to weights of the student network in the last epoch. The loss function of PS-KD can be formulated as follows:
\begin{equation}
\label{eq:pskd}
    \mathcal{L}_{\mathrm{PS}-\mathrm{KD}}=\mathcal{L}_{SD}\left(x_{i}; \theta_{s}, \bar{\theta}_{s}; \alpha\right),
\end{equation}
where $\bar{\theta}_{s}$ represents the weights of the network in the last epoch, and $\alpha$ is an additional coefficient that controls the proportion of historical knowledge from the previous epoch.
Different from the original settings in PS-KD, we modified the \textit{epoch-level} distillation to the \textit{iteration-level} for the pruning in EPSD. Namely, EPSD utilized each iteration samples $x_{i}$ to generate the predictions, and distilled the predictions as soft targets in the next iteration. This modification avoids traversing the entire training dataset in pruning, making our method more efficient in the pruning phase. Besides, we set $\alpha$ to a small fixed value when pruning, considering that the network usually does not have enough knowledge about data in the early stages of training. The detailed hyperparameters can be found in Table~\ref{tab:imp_pskd}.

\noindent\textbf{Impact of Hyper-parameter $\mathbf{\alpha}$}. Historical information plays an important role when EPSD is equipped with PS-KD for generating soft targets, which is controlled by the hyperparameter $\alpha$. 
During training, we follow the original paper's method of gradually increasing $\alpha$ with training epochs. For pruning, we opt for a constant $\alpha$ value for simplicity and explore this fixed value alongside PS-KD pruning experiments. Table~\ref{tab:abl_cifar_alpha} outlines two scenarios with different $\alpha$ values for VGG-16 and ResNet-18 on CIFAR-10 and CIFAR-100 datasets. $\alpha$ varies between 0.1 and 0.4, and we report the top-1 accuracy. Sparsity remains at 80\% across all setups.
As illustrated in Table~\ref{tab:abl_cifar_alpha}, EPSD achieves the higher accuracy  with $\alpha=0.1$ using VGG16 on CIFAR-10, and $\alpha=0.2$ for ResNet-18 on CIFAR-100. We observe that when $\alpha$ changes from 0.1 to 0.4, the difference between the upper and lower accuracy bounds of the two settings is stable within 0.3\% (94.44\%-94.15\%) and 0.5\% (79.44\%-78.96\%), respectively. Meanwhile, EPSD (80\% sparsity) still shows better results than the two unpruned baselines in all configurations. The above analysis shows that our method is not sensitive to the hyperparameters $\alpha$ of PS-KD.

\noindent \textbf{More Results}. We provided more results of EPSD equipped with PS-KD to verify the results in the main manuscript, as shown in Fig.~\ref{fig:supp_main_results} (second column). It can be found that EPSD consistently outperformed the `Simple Combination' over all settings, and EPSD outperformed `Unpruned Baseline' and `SD Only' in most settings, indicating that pruning can boost the performance of SD. For instance, with ResNet-18 on CIFAR-100 at sparsity 90\%, the accuracy of EPSD is 77.33\%, which is 2.04\% higher than the `Unpruned Baseline'. With ResNet-18 on Tiny-ImageNet at sparsity 36\%, the accuracy of EPSD is 0.93\% higher than the `SD Only' (57.76\% vs. 56.83\%).

\subsection{EPSD equipped with DLB}
\noindent \textbf{Implementations}. For DLB~\cite{shen2022self}, Shen et al. introduce an extra last-batch consistency regularization loss. Rather than storing the whole predictions at the last iteration as designed in PS-KD, DLB employs a data sampler to obtain batches $\mathcal{B}_t$ and $\mathcal{B}_{t-1}$ in iteration t and t-1 simultaneously at the (t-1)$^{th}$ iteration for implementation. Whereas predictions from $\mathcal{B}_t$ are smoothed by temperature $\tau$ and then stored for regularization in t$^{th}$ iteration. The overall loss function is formulated by:
\begin{equation}
\label{eq:dlb}
    \mathcal{L}_{\mathrm{DLB}}=\mathcal{L}_{\mathrm{CE}}(x_{i}; \theta_{s})+\lambda_{cls} \cdot \mathcal{L}_{SD}\left(x_{i}; \theta_{s}, \bar{\theta}_{s}; \tau\right),
\end{equation}
where $\lambda_{cls}$ is the coefficient to balance two loss terms. The definition of symbols $\bar{x}_{i}$ and $\bar{\theta}_{s}$ for Eq.~(\ref{eq:supp_sd_loss}) in DLB is replaced with the identical input $x_{i}$ and the weights of student network at last \textit{iteration}, respectively. DLB divides each batch into two halves: one aligns with the previous iteration, and the other with the next iteration. The first half batch distills using real-time softened targets from the previous iteration. Specific hyperparameters are provided in Table~\ref{tab:imp_dlb}.

\noindent\textbf{More Results}. We provide more results of EPSD combined with DLB on different network structures and datasets to supplement the results in the main manuscript, as shown in Fig.~\ref{fig:supp_main_results} (3rd column). It can be found that EPSD kept comparable accuracies under high sparsity (e.g., 95\%) while the accuracies of the `Simple Combination' heavily decreased. For instance, with ResNet-18 on CIFAR-100 at sparsity 95\%, EPSD achieved 72.72\% accuracy while the accuracy of the `Simple Combination' was only 46.86\%.

\section{Ablation Study}
\label{sec:appendix_abl}
\begin{table}
\centering
\small
\renewcommand\arraystretch{1.25}
\setlength{\tabcolsep}{2mm}{
\begin{tabular}{cc|cc|c}
    \toprule
      \multicolumn{2}{c|}{\textbf{Step1: Pruning w/}} & \multicolumn{2}{c|}{\textbf{Step2: Training w/}} & \multirow{2}{*}{Test Acc. (\textit{$\%$})} \\
      \textit{CE Loss} & \textit{SD Loss} & \textit{CE Loss} & \textit{SD Loss} & \\ \midrule
      None & None & \checkmark & - & $78.88$ \\ \midrule
      \checkmark & - & \checkmark & - & $76.80_{(\textcolor[RGB]{0,255,0}{\downarrow2.08})}$ \\
      - & \checkmark & \checkmark & - & $77.85_{(\textcolor[RGB]{0,255,0}{\downarrow1.03})}$ \\
      \checkmark & - & - & \checkmark & $79.30_{(\textcolor[RGB]{255,0,0}{\uparrow0.42})}$ \\
      - & \checkmark & - & \checkmark & $\mathbf{79.85}_{(\textcolor[RGB]{255,0,0}{\uparrow0.97})}$ \\
    \bottomrule
\end{tabular}
}
\caption{Ablation study for EPSD using ResNet-18 on CIFAR-100 dataset. 
The hyperparameters are kept consistent for pruning and training, and the numerical subscripts indicate the percentage of performance increase or decrease relative to the unpruned baseline (\textit{first} row).}
\label{tab:abl_c100_all}
\end{table}
We unveil the efficacy of EPSD by varying the optimization objectives (standard cross-entropy loss vs. SD loss) in two steps. Moreover, we investigate the impact of employing another early pruning method SNIP~\cite{lee2018snip} on EPSD.

\subsection{Varying Objectives}
To show the performance improvement of the different objectives in step-1 and step-2 mentioned in the manuscript, we use SD loss and cross-entropy (CE) loss to differentiate the influence of each component in the two steps of EPSD. The ResNet-18 on CIFAR-100 is adopted as the study case. Specifically, first, we fixedly use CE loss in step-2 to train the pruned network and use SD loss or CE loss to evaluate the importance of weights in the pruning of step-1, to observe which loss is beneficial to retain distillable weights. Then, we fixedly use SD loss for optimization in step-2, and use these two losses for pruning, to observe which loss is more conducive to retaining distillable weights.
\textbf{Step-1: Early pruning with/without SD}. As shown in Table~\ref{tab:abl_c100_all}, when step-2 is trained with CE loss (1st, 2nd row), the performance of the pruned network decreased compared with the unpruned baseline (78.88\%), and the degradation is less by pruning with SD loss (-1.03\%) compared with pruning with CE loss (-2.08\%). This showed that pruning with SD can preserve more trainable weights. On the other hand, when step-2 is trained with SD loss (3rd, 4th row), pruning with SD loss (79.85\%) is still better than pruning with CE loss (79.30\%).
\textbf{Step-2: Training with/without SD}. We further investigated the effectiveness of the SD, which trains the pruned sub-network in EPSD. In Table~\ref{tab:abl_c100_all}, the fourth and fifth rows show that, when we use the SD loss instead of the CE loss, training with SD can achieve better results than the unpruned baseline. Specifically, compared to training with CE, training with SD improves by about 2.5\%, and pruning with soft-gradient is also improved by about 2\%. In addition, with the help of SD, they can even surpass the unpruned network by 0.42\% and 0.97\%, respectively.
The above analysis shows that the two improvements both achieve the expected improvement, which further proves the efficiency of our proposed model compression method.

\subsection{Effect of Pruning algorithm}
\label{sec:supp_snip}
\begin{figure}[!t]
\centering
\includegraphics[width=1.0\linewidth]{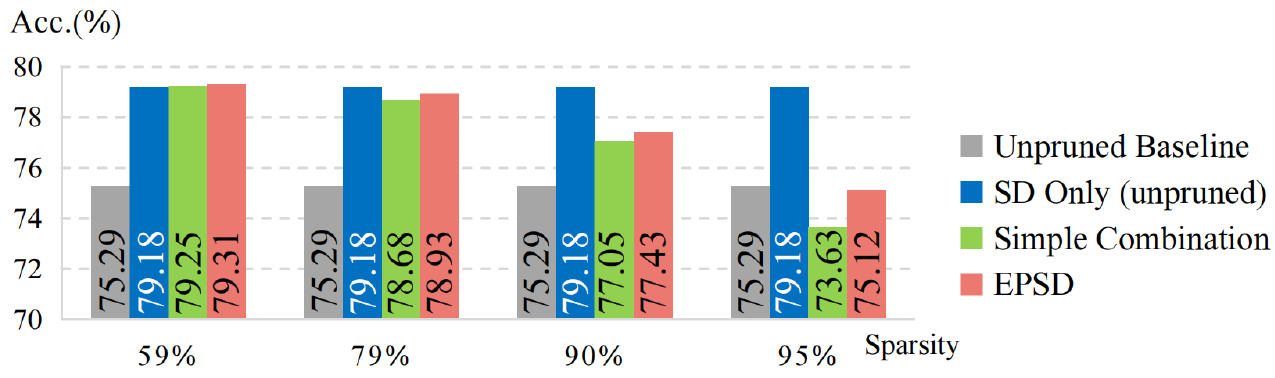}
\caption{Combining SNIP and PS-KD with ResNet-18 on CIFAR-100. We compare the simple combination of SNIP and SD (`Simple Combination'), EPSD, and pre-trained network (`Unpruned Baseline'), and the network only performs PS-KD without any sparsity (`SD Only'). }
\label{fig:supp_snip}
\end{figure}
We applied another early pruning algorithm, SNIP~\cite{lee2018snip}, to our proposed framework EPSD. The SNIP is a simple pruning algorithm that only considers the immediate impact of pruning on the loss before training. We combined SNIP with PS-KD as a study case to investigate the performance. 
As shown in Fig.~\ref{fig:supp_snip}, our framework still consistently outperformed the `Simple Combination' over all settings when using SNIP for pruning. For instance, under high sparsity conditions (e.g. 95\%), the accuracy of the `Simple Combination' is 1.66\% lower than the `Unpruned Baseline' while EPSD is comparable (75.12\% vs. 75.29\%). Our results indicate that EPSD's effectiveness is independent of a particular pruning algorithm.



\end{document}